\documentclass[preprint, 10pt]{elsarticle}

\usepackage[margin=2.5cm]{geometry}
\usepackage{setspace}
\usepackage{lmodern}
\usepackage{amsmath,amssymb}
\usepackage{amsfonts,amsthm}
\usepackage{lineno}
\usepackage{graphicx}
\usepackage{bm, bbm}
\usepackage{algorithm}
\usepackage{physics}
\usepackage{color}
\usepackage{pxfonts}
\usepackage[footnotesize,bf]{caption}
\usepackage{subcaption}
\usepackage{mathtools}
\usepackage{hhline}
\usepackage[T1]{fontenc}

\usepackage{placeins}

\newcommand{\R}{\mathbb{R}}

\newtheorem{remark}{Remark}[section]

\graphicspath{{figures/}}
\journal{arXiv}

\begin{document}
\begin{frontmatter}
\title{Normalizing field flows: Solving forward and inverse stochastic differential equations using physics-informed flow models}
\author[SNH]{Ling Guo}
\author[TJ]{Hao Wu}
\author[LESC]{Tao Zhou}
\address[SNH]{Department of Mathematics, Shanghai Normal University, Shanghai, China}
\address[TJ]{School odmathematical sciences, Tongji University, Shanghai, China}
\address[LESC]{LSEC, Institute of Computational Mathematics and Scientific/Engineering
Computing, Academy of Mathematics and Systems Science, Chinese Academy
of Sciences, Beijing, China.}

\begin{abstract}
We introduce in this work the \textit{normalizing field flows} (NFF) for learning random fields from scattered measurements. More precisely, we construct a bijective  transformation (a normalizing flow characterizing by neural networks) between a Gaussian random field with the Karhunen-Lo\`eve (KL) expansion structure and the target stochastic field, where the KL expansion coefficients and the invertible networks are trained by maximizing the sum of the log-likelihood on scattered measurements. This NFF model can be used to solve data-driven forward, inverse, and mixed forward/inverse stochastic partial differential equations in a unified framework. We demonstrate the capability of the proposed NFF model for learning Non Gaussian processes and different types of stochastic partial differential equations.
\end{abstract}

\begin{keyword}
Data-driven modeling \sep normalizing flows \sep uncertainty quantification \sep random fields
\end{keyword}

\end{frontmatter}

\section{Introduction}\label{S:1}
Inherent uncertainties arise naturally in complex engineering systems due to model errors,  stochastic excitation, unknown material properties or random initial/boundary conditions. This yields an active research area of  uncertainty quantification (UQ). The main purpose of UQ is to quantify the effect of these various sources of randomness on output model predictions. Such problems are usually reformulated as mathematical models with high-dimensional random inputs. Classical approaches in the field of UQ usually rely on model order reduction techniques, such as the Karhunen-Lo\`eve (KL) expansion, to represent the infinite input random field into finite low dimensional parameter variables and then construct surrogate models of the quantities of interests (QoI) in the parametric space, and one can then obtain the statistic information of the QoI. For more details, one can refer to \cite{Xiu_2002Wiener,billions16,guoReview2020} and references therein. Such classical approaches in general suffer from the curse of dimensionality. Moreover, such approaches also require prior knowledge of the input randomness (such as the covariance function) which is not practical for many applications. For example, in some cases one only has small amount of measurements of the stochastic input, which leads to data driven problems.

In recent years, machine learning techniques have been widely investigated to deal with data driven forward and inverse partial differential equations. Popular machine learning tools include Gaussian process \cite{graepel2003,sarkka2011,Raissi_nonlinear} and deep neural networks (DNNs) \cite{Lagaris1997,Lagaris2000,Yinglexing17,MaziarParisGK17_1,Stuart10,zhu_zabaras18,Rudy17,MaziarParisGK17_2,MaziarGK18JCP,Tartakovsky2018}. Inspired by these developments on learning deterministic PDEs, there have also been recent advances on using deep learning techniques to solve stochastic differential equations. For example, a deep learning approximation for high-dimensional SDEs is proposed in \cite{Weinan-arxiv} for forward stochastic problems; Zhang et.al. \cite{dongkunNNapc2018} have recently proposed a DNN based arbitrary polynomial expansion method that learns the modal functions of the quantity of interest for SDEs with random inputs. Although it is a unified framework for both forward and inverse problems, it still suffers from the "curse of dimensionality" in that the number of polynomial chaos expansion terms grows exponentially as the effective dimension increases. In \cite{yangliusisc2020}, physics-informed generative adversarial models (PI-GAN) is developed to possibly tackle SDEs involving stochastic processes with high effective dimensions. 

Recently, flow-based models have also been successfully applied to various real applications including computer vision \cite{Kingma2018,Ho2019}, audio generation \cite{Esling2019,Kim2018} and physical systems \cite{Koller2009,Noe2019}. A normalizing flow (NF) provides an exact log-likelihood evaluation and thus can produce tractable distributions where both sampling and density evaluation can be efficient and exact. We refer to \cite{Kobyzev2020} for more details on comprehensive review of the NF. In particular, there have been great interests in learning complex stochastic models with flow-based generative models. For example, BayesFlow is proposed in \cite{bayesflow2020} for globally amortized Bayesian inference. Padmanabha et al.\cite{zabaras2021} developed
efficient surrogates using conditional invertible neural networks to recover a high-dimensional non-Gaussian log-permeability field in multiphase flow problem.

In this work, we aim at proposing the normalizing field flow (NFF) model for solving forward and inverse stochastic problems. In our setting, we assume that we have partial information of the stochastic input (such as the diffusivity/forcing terms in the stochastic problems) or the solution, from scattered sensors. Thus, the type of the data-driven problem varies depending on the information we have: a forward problem if partial information of the stochastic input is given and an inverse problem when partial information of the solution is available. Our goal is to infer the unknown stochastic fields, and quantify their uncertainties given the randomness in the data.  To this end, we build the NFF model for the unknown stochastic field in three steps: 1) we construct a reference Gaussian random filed $z(x,\omega)$ with a truncated Karhunen-Lo\`eve (KL) expansion structure, where the expansion coefficients are parameterized by deep neural networks; 2) we then construct a bijective transformation between the reference field $z(x,\omega)$ and the target stochastic field by a normalizing flow; 3) all the parameters involved are then trained by maximizing the sum of the log-likelihood on the observable measurements. Moreover, for SDE problems, we encode the known physics and include the SDE loss into the total loss, which results in the physics informed flow model. Our approach admits the following main advantages:
\begin{itemize}
\item Our NFF model is fully data-driven, and it can be used to solve forward, inverse, and mixed stochastic partial differential equations in a unified framework.
\item Unlike the setting of existing DNN based Non-Gaussian random field models (e.g., \cite{yangliusisc2020}), sensor locations are not assumed to be fixed for different snapshots in our flow model.
\item The physics informed NFF model is adopted to learn random fields in UQ problems, and it can alleviate the \textit{curse of dimensionality} that appears in  most traditional approaches (such as polynomial chaos).
\end{itemize}
Several numerical tests are presented to illustrate the effectiveness of the new NFF method.

The organization of this paper is as follows. In Section \ref{S:2}, we set up the data-driven forward and inverse problems. In Section \ref{S:3}, we introduce the flow-based generative models, followed by our main algorithm -- the normalizing field flow method (NFF) for solving SDEs. In Section \ref{S:4}, we first present a detailed study of the accuracy and performance of our NFF model for leaning stochastic processes including Non-Gaussian and mixed Non-Gaussian fields. Then we present the simulation results for solving both the forward and inverse stochastic elliptic equations. Finally, we give some concluding remarks in Section \ref{S:5}.

\section{Problem Setup}\label{S:2}
Let $(\Omega,F,P)$ be a probability space, where $\Omega$ is the sample space, $F$ is the $\sigma$-algebra of subsets of $\Omega$, and $P$ is a probability measure. We consider the following stochastic partial differential equation:
\begin{equation}\label{eqn:SDE}
\begin{gathered}
    \mathcal{N}_x[u(x;\omega);k(x;\omega)] = f(x;\omega), \quad x\in\mathcal{D}, \quad \omega\in\Omega,\\
    \mathcal{B}_x[u(x;\omega)] = 0, \quad x\in\Gamma,
\end{gathered}
\end{equation}
where $\mathcal{N}_x$ is the general form of a differential operator that could be nonlinear, $\mathcal{D}$ is the $d$-dimensional physical domain in $\R^d$ ($d = 1,2,$ or 3), and $u(x;\omega)$ is the exact solution. The boundary condition is imposed through the generalized boundary condition operator $\mathcal{B}_x$ on $\Gamma$. Here $k(x;\omega)$ and $f(x;\omega)$ could be the sources of uncertainty, which can be represented by random fields.

In this work, we consider the scenario that we have limited number of scattered sensors to generate measurements for the stochastic fields. We denote the measurements from all the sensors at the same instant by a {\em snapshot}, and assume that every snapshot of sensor data corresponds to the same random event in the random space. We also assume that when the number of snapshots is big enough, the empirical distribution approximates the true distribution.

Suppose there are a total number of $N_s$ snapshots of measurements of  $k(x;\omega)$, $u(x;\omega)$ and $f(x;\omega)$ observed from the sensors. For each snapshot $s$,
we have $N_k$, $N_f$, $N_u$ sensors for $k,f,u$ placed at $\{x_{k,i}^{s}\}_{i=1}^{N_k}$, $\{x_{f,i}^{s}\}_{i=1}^{N_f}$, $\{x_{u,i}^{s}\}_{i=1}^{N_u}$ respectively. Let $k_i^{s}$,$f_i^{s}$ and $u_i^{s}$ ($s=1,2,...,N_s$) be the $s$-th measurement of $k$, $f$ and $u$ at location $x_{k,i}^{s}$, $x_{f,i}^{s}$ and $x_{u,i}^{s}$ respectively, and $\omega^s$ is the random instance at the $s$-th measurement, i.e., $k_i^{s} = k(x_{k,i}^{s};\omega^s)$, $f_i^{s} = f(x_{f,i}^{s};\omega^s)$ and $u_i^{s} = u(x_{u,i}^{s};\omega^s)$. Then the accessible data set can be represented by
\begin{equation}\label{eqn:training_set}
    \mathcal{S}_t = \left\{\{(x_{k,i}^s, k_i^s)\}_{i=1}^{N_k}, \{(x_{u,i}^s, u_i^s)\}_{i=1}^{N_u}, \{(x_{f,i}^s, f_i^s)\}_{i=1}^{N_f}\right\}_{s=1}^{N_s}.
\end{equation}
We always assume that a sufficient number of sensors are given for $f(x;\omega)$. Thus the type of the data-driven problem (\ref{eqn:SDE}) varies depending on what information we have for $k(x;\omega)$ and $u(x;\omega)$. The problem will transform from forward problem to inverse problem as we decrease the number of sensors $N_k$ on $k(x;\omega)$ while increase the number of sensors $N_u$ on $u(x;\omega)$.

\begin{remark}
Unlike the settings often considered in existing DNN based random field models (e.g., \cite{yangliusisc2020}), sensor locations are not assumed to be fixed for different snapshots in this paper. See more details in Section 4.
\end{remark}

\section{Methodology}
\label{S:3}
\subsection{Normalizing flow with composing invertible networks}
\label{S:3-1}
In this part, we will first briefly review normalizing flows (in particular, the coupling flows) for distribution learning. The basic idea of a normalizing flow is to learn a complex distribution (the target distribution) by a transformation from a simple distribution (the reference distribution) via a sequence of invertible and differentiable mappings. To illustrate the idea, let $Z\in \R^M$ be a random variable and its probability density function $p_Z(z)$ is given (e.g. Gaussian distribution). Let $\mathcal{F}_{ZK}$ be a bijective mapping and $K=\mathcal{F}_{ZK}(Z)$. Then we can generate $K$ samples through $p_Z(z)$. Furthermore, we can compute the probability density function of the random variable $K$ by using the change of variables formula:
\begin{equation}\label{eqn:change_variable}
p_K(k)=\bigg |\text{det} \frac{\partial \mathcal{F}_{KZ}(k)}{\partial k} \bigg | p_Z(z),
\end{equation}
where $\mathcal{F}_{KZ}$ is the inverse map of $\mathcal{F}_{ZK}$. Assume the map $\mathcal{F}_{ZK}$ is parameterized by $\theta$ and the reference measure $p_Z(z)$
 (or, the bases measure), is parameterized by $\phi$. Given a set of observed training data $\mathcal{D}=\{k_i\}_{i=1}^{N}$ from $k$, the log likelihood is given as following by taking log operator on both sides of (\ref{eqn:change_variable}):
\begin{equation}\label{eqn:likelihood1}
    \begin{aligned}
    \text{log}~p_K(\mathcal{D}|\theta,\phi) &=\sum_{i=1}^{N}\text{log}~p_K(k_i|\theta,\phi)\\
    & = \sum_{i=1}^{N}\text{log}~p_Z(z_i|\phi)-\text{log}~\bigg |\text{det} \frac{\partial \mathcal{F}_{ZK}(z_i)}{\partial z} \bigg |.
    \end{aligned}
\end{equation}
The parameters $\theta$ and $\phi$ can be optimized during the training by maximizing the above log-likelihood.

Notice that invertible and differentiable transformations are composable. Thus in practice we can chain together multiple bijective functions $\mathcal{F}_{1},\cdot\cdot\cdot,\mathcal{F}_{n}$ to obtain
$\mathcal{F}_{ZK}=\mathcal{F}_n\circ \mathcal{F}_{n-1}\circ \cdot\circ \mathcal{F}_1$, which can approximate more complex distribution $p_K(k)$.  Denote the intermediate
variables by $Z^{n-1},...,Z^1,Z^0$, then the log-likelihood of the
complex distribution $p_K(k)$ can be expressed as
\begin{equation}\label{eqn:likelihood2}
    \text{log}~p_K(k) =\text{log}~p_{Z^0}(z^0)-\sum_{j=1}^{n}\text{log}~\bigg|\text{det}\frac{\partial \mathcal{F}_j(z^{j-1})}{\partial z^{j-1}}\bigg|.
\end{equation}

\begin{figure}[htbp]
    \centering
    \begin{subfigure}{0.45\textwidth}
        \centering
        \includegraphics[width=\linewidth]{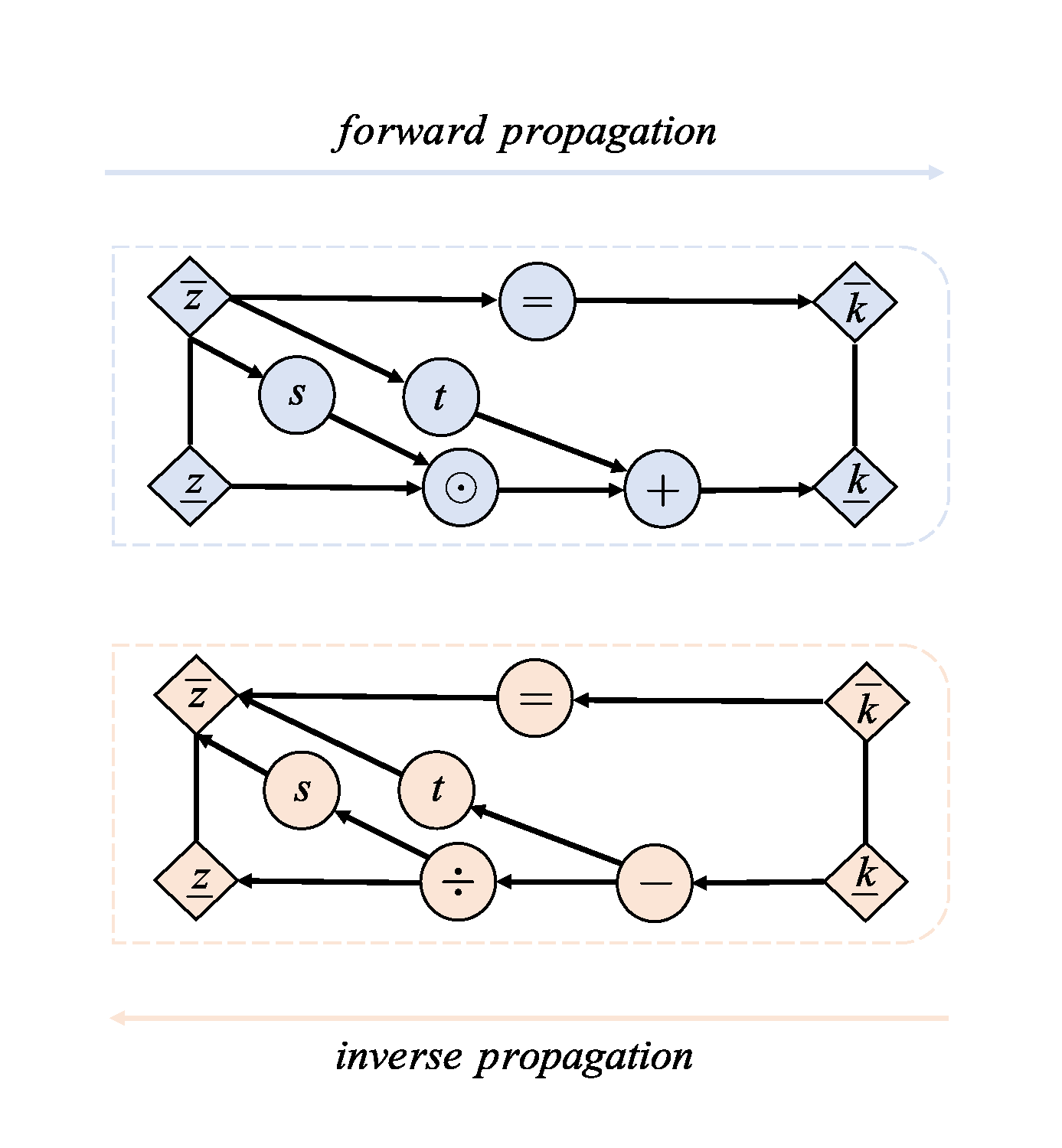}
        \caption{}\label{fig:realnvp}
    \end{subfigure}%
    \begin{subfigure}{0.45\textwidth}
        \centering
        \includegraphics[width=\linewidth]{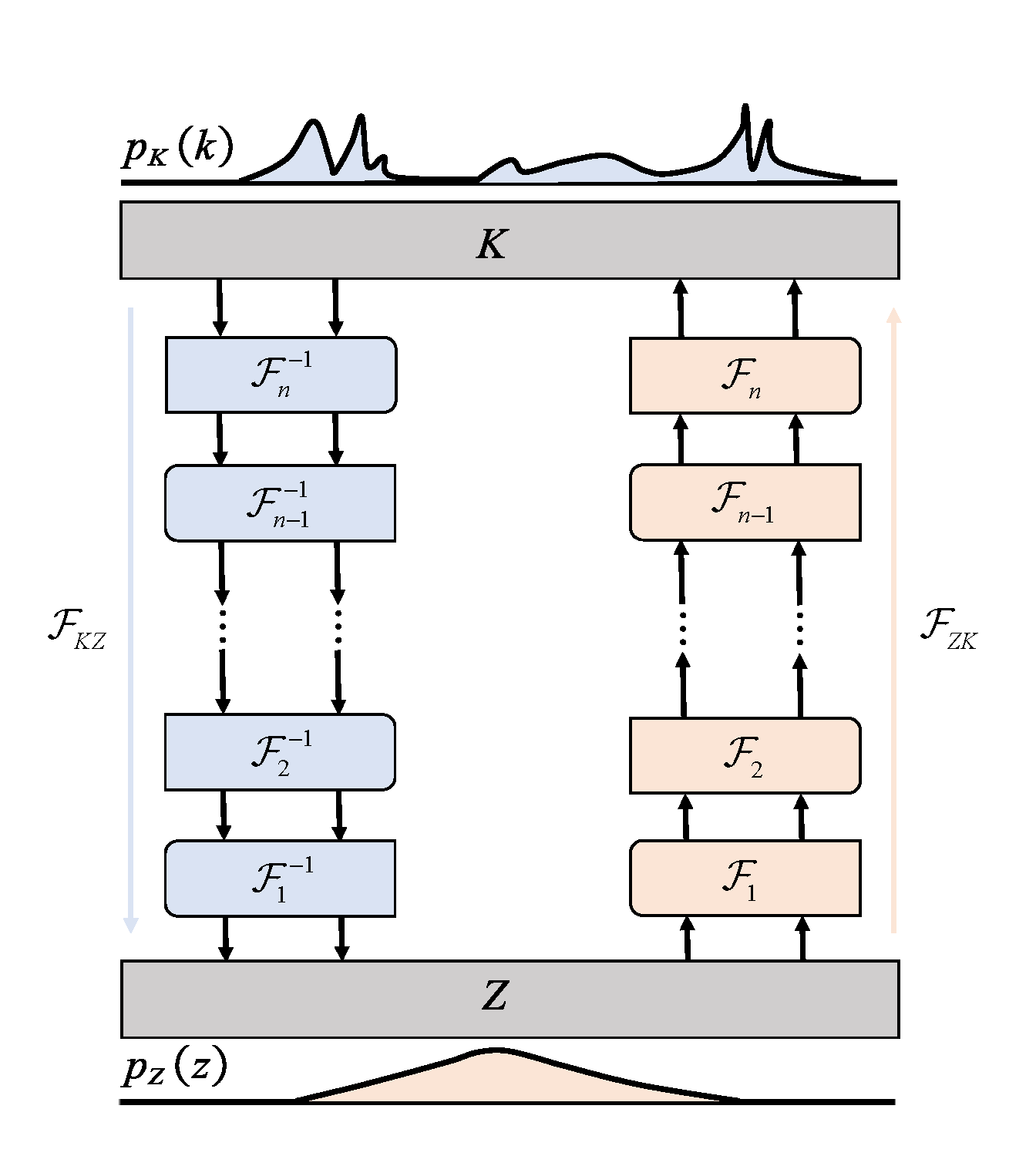}
        \caption{}
    \end{subfigure}
    \caption{(a):Illustration of realNVP model; (b): Normalizing flow with $n$ coupling layers $\mathcal{F}$. }\label{fig:INN_ZK}
\end{figure}

Many different types of flow maps have been constructed and investigated in literature, we refer to ~\cite{Kingma2018,Dinh2015,Dinh2017} and a recent review paper \cite{Kobyzev2020} for more details. In this work, we will adopt the real-valued Non-Volume Preserving (RealNVP) model proposed by Dinh et al. in ~\cite{Dinh2017} as our basic block to construct the invertible transformation $\mathcal{F}_{ZK}$. Specifically, the input vector $z\in \R^M$ is split into two parts, denoted by $\bar{z}\in \R^m$ and $\underline {z}\in \R^{M-m}$ respectively. Then one can define a transformation
block $\mathcal{F}_i$
by the following formula
\begin{equation}\label{realnvp1}
\begin{array}{cc}
   \bar{k}=\bar{z}  &  \underline {k}=\underline {z}\odot \text{exp}(s(\bar{z}))+t(\bar{z}), \quad \text{if $i$ is odd},\\
   \underline {k}= \underline{z}  & \bar{k}=\bar {z}\odot \text{exp}(s(\underline {z}))+t(\underline {z}), \quad \text{if $i$ is even},
\end{array}
\end{equation}
where $\odot$ represents the element-wise multiplication, $s(\cdot)$ and $t(\cdot)$ are realized by fully connected neural networks in our implementation(see Figure (\ref{fig:realnvp})).
Following ~\cite{Dinh2017}, for the odd $i$ the Jacobian determinant of $\mathcal{F}_i$ is given by
\begin{equation}
\mathcal{J}=
\left[
  \begin{array}{cc}
    \mathbb{I}_m & \mathbf{0}_{m\times(M-m)} \\
    \frac{\partial \underline {k}}{\partial \bar{z}} & \text{diag}~(\text{exp}(s(\bar{z}))) \\
  \end{array}
\right],
\end{equation}
and the inverse transformation block can be computed directly (see Fig.\ref{fig:INN_ZK}),
\begin{equation}\label{realnvp_inverse}
    \bar{z}=\bar{k}, \quad \quad \underline {z}=(\underline {k}-t(\bar{k}))\odot \text{exp}(-s(\bar{k})).
\end{equation}
For the even $i$, the Jacobian determinant of $\mathcal{F}_i$ and the inverse can be calculated in the same way.

\subsection{NFF: Normalizing field flows}
\label{S:3-2}
We are ready to present our normalizing field flows models. To this end, we first introduce flow based generative model for approximating a stochastic vector field $k=k(x;\omega)\in\mathbb{R}^{D}$ defined on $\mathbb{R}^{D_{x}}\times \Omega$. We use the total $N_s$ snapshots as our training data, where each snapshot $s$ contains measurements
\begin{equation}
   \{k^s_{1:N}\}_{s=1}^{N_s} = \{(k^s_1,\ldots,k^s_{N})\}_{s=1}^{N_s} =\{(k(x_i^s,\omega^s))_{i=1}^{N}\}_{s=1}^{N_s}
\end{equation}
collected at $x_{1:N}^s=(x_{1}^s,\ldots,x_{N}^s)$. Here we omit the subscript $k$ of $N_k$ and $x_{k,i}^s$ for simplicity unless confusion arises.
Inspired by the idea of the normalizing flow, we can model the conditional distribution of random variable $k$ given $x$ as $K=\mathcal{F}_{ZK}(Z,x)$, the conditional probability density function is given by:
\begin{equation}\label{eqn:change_variable_conditional}
p_K(k|x)=\bigg |\text{det} \frac{\partial \mathcal{F}_{KZ}(k,x)}{\partial k} \bigg |~ p_Z(z|x).
\end{equation}

Then the main goal of the normalizing field flows is to model the conditional distribution $\mathbb P(k(x,\cdot)|x)$ with the training data via the invertible map $\mathcal{F}_{ZK}(Z,x)$. Concretely, NFF consists of two steps:
\begin{enumerate}
    \item Construct a reference Gaussian random filed $z(x,\omega)$ as
\begin{eqnarray}\label{eqn:INNmodel1}
z(x,\omega) & = & A(x)+ B(x)\xi(\omega)+\mathrm{diag}\left(C(x)\right)\epsilon(x,\omega).
\end{eqnarray}
where $A(x)\in\mathbb{R}^{D}$, $B(x)\in\mathbb{R}^{D\times M}$ and $C(x)\in\mathbb{R}^{D}$ are taken as fully connected neural networks. $\xi(\omega)\in\mathbb{R}^{M}  \sim  \mathcal{N}(0,\mathbf{I_M})$
and $M$ denotes the order of the finite truncation terms of the KL expansion. To avoid the singularity of the covariance matrix, we add the last term and
assume $\epsilon(x,\omega)  \stackrel{\mathrm{iid}}{\sim}  \mathcal{N}(0,\mathbf{I_D})$ is a
Gaussian noise.

\item Build a bijective transformation given $x$ between the target random field $k(x,\omega)$ and reference field $z(x,\omega)$:
\begin{equation}\label{eqn:INNmodel}
k(x,\omega)=\mathcal{F}_{ZK}(z(x,\omega),x),
\end{equation}
Where $\mathcal{F}_{ZK}$ ia s normalizing flow.
\end{enumerate}
\begin{figure}[htbp]
	\centering
	\includegraphics[width=0.45\linewidth]{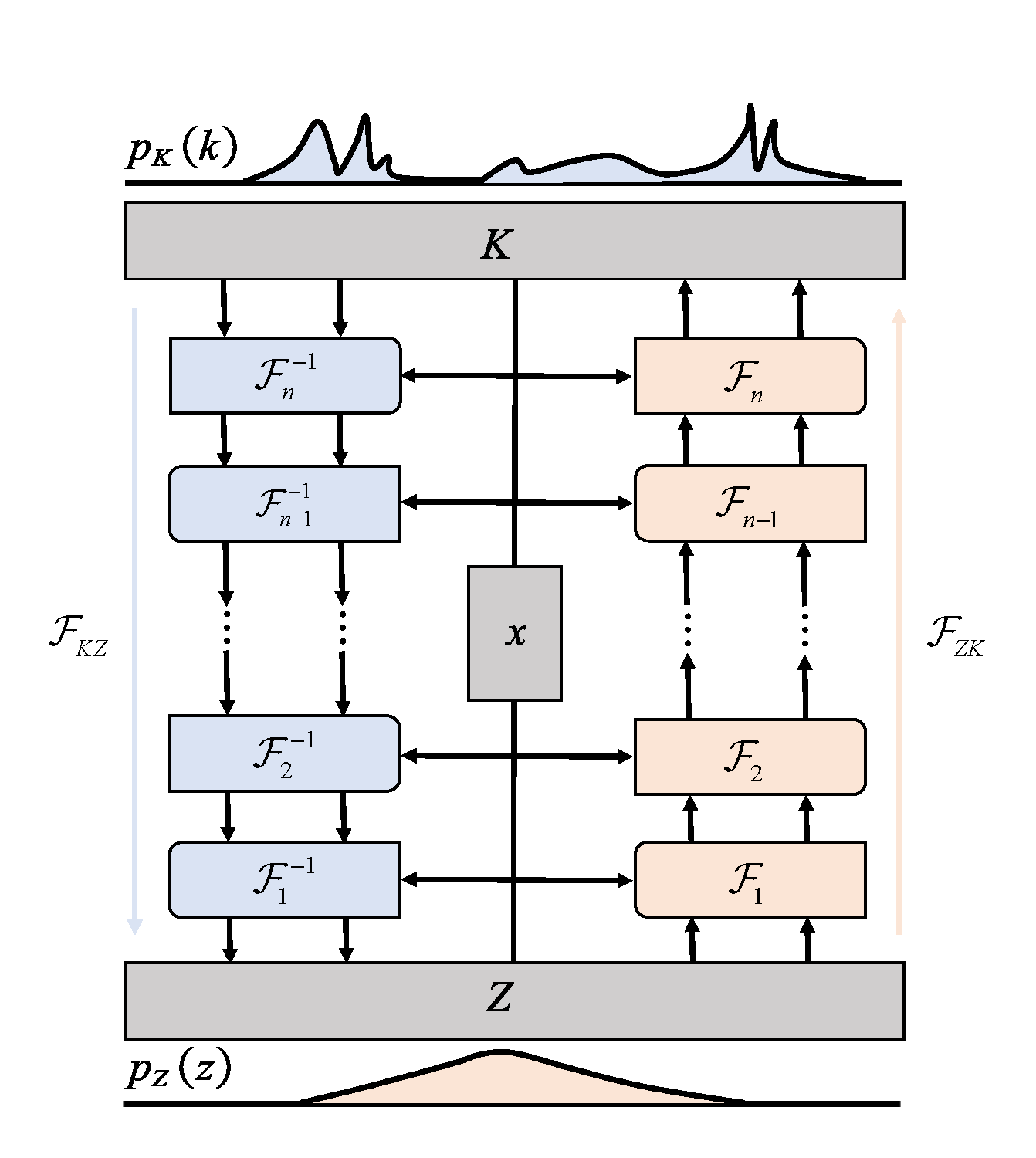}
	\caption{Schematic of normalizing field flow based on realNVP.}
	\label{fig:NFF_ZK_x}
\end{figure}
The detail schematic of NFF model is shown in Fig.~\ref{fig:NFF_ZK_x}. Given sensor locations $x_{1:N}^s$ of snapshot $s$, it is straight forward to obtain from (\ref{eqn:INNmodel1}) that $z_{1:N}^s=(z(x_1^s,\omega^s),\cdot\cdot\cdot,z(x_N^s,\omega^s))^T\sim \mathcal{N}(\boldsymbol{\mu^s},\boldsymbol{\Sigma^s})$ with
\begin{equation}\label{eqn:statistics}
\boldsymbol{\mu}^s=\left(\begin{array}{c}A(x_1^s) \\ \vdots \\A(x_N^s)\end{array}\right),\quad  \boldsymbol{\Sigma}^s=\left(\begin{array}{c}B(x_1^s) \\ \vdots \\B(x_N^s)\end{array}\right)\left(\begin{array}{c}B(x_1^s) \\ \vdots \\B(x_N^s)\end{array}\right)^{T}+\text{diag}\left(\begin{array}{c}C(x_1^s) \\ \vdots \\C(x_N^s)\end{array}\right)^2.
\end{equation}
Our goal is to train neural networks $A(x),B(x),C(x)$ and the invertible neural network $\mathcal{F}_{ZK}$ to minimize the likelihood objective
\begin{equation}\label{eqn:INNmodelloss}
\mathcal{L_{\text{data}}}  =  -\frac{1}{N_s}\sum\limits_{s=1}^{N_s}\log \mathbb{P} (k^s_{1:N}|x^s_{1:N}),
\end{equation}
where
\begin{eqnarray}\label{eqn:INNmodellikelihood}
\log\mathbb{P}(k_{1:N}^{s}|x_{1:N}^{s}) & = & \log\mathbb{P}(z_{1:N}^{s}|x_{1:N}^{s})+\sum_{i=1}^{N}\log\left|\mathrm{det}\frac{\partial F_{KZ}(k_{i}^{s},x_{i}^{s})}{\partial k_{i}^{s}}\right|\nonumber\\
 & = & \log\mathcal{N}(z_{1:N}^{s}|\boldsymbol{\mu}^{s},\boldsymbol{\mathbf{\Sigma}}^{s})+\sum_{i=1}^{N}\log\left|\mathrm{det}\frac{\partial F_{KZ}(k_{i}^{s},x_{i}^{s})}{\partial k_{i}^{s}}\right|.
\end{eqnarray}

\begin{remark}
We can of course construct a more specific reference field, i.e., $A(x),B(x),C(x)$ can be of explicit formulas. However, we model them there by neural networks to enhance the expression capacity of the filed, and moreover, we aim at training the reference filed by data. We remark that the optimal way for finding the reference filed is an open question.
\end{remark}

\begin{remark}
In the case where $N\gg M$, $\boldsymbol\Sigma^s$
can be decomposed into a low-rank matrix and a diagonal matrix as in \eqref{eqn:statistics},
and the log-likelihood $\log\mathcal{N}(\cdot|\boldsymbol{\mu}^{s},\boldsymbol{\mathbf{\Sigma}}^{s})$ can be efficiently obtained by calculating the determinant and inverse of a $M$-by-$M$ sized matrix according to the Woodbury matrix identity \cite{higham2002accuracy} and the matrix determinant lemma \cite{harville1998matrix}. In our experiments, we implement the fast algorithm by the class \texttt{LowRankMultivariateNormal} in PyTorch \cite{paszke2017automatic}.
\end{remark}

Once the NFF has been constructed, for a set of new sensor measurements $k_{1:n}=\{k(x_i,\omega)\}_{i=1}^n$ at locations
$x_{1:n}$, we can predict values of $k(x,\omega)$ at other $n'$ locations $x_{n+1:n+n'}$  by drawing samples from their conditional distribution
 \begin{eqnarray}
\mathbb{P}(k_{n+1:n+n'}|k_{1:n},x_{1:n+n'}) & = & \int\mathbb{P}(k_{n+1:n+n'}|\xi,x_{n+1:n+n'})\mathbb{P}(\xi|k_{1:n},x_{1:n})\mathrm{d}\xi,\nonumber\\
 & = & \int\mathbb{P}(k_{n+1:n+n'}|\xi,x_{n+1:n+n'})\mathbb{P}(\xi|z_{1:n},x_{1:n})\mathrm{d}\xi.
\end{eqnarray}
Since $z(x,\omega)$ is a Gaussian random field, the posterior distribution of $\xi$ for given $z_{1:n}$ and $x_{1:n}$
is a multivariate distribution, i.e.,
\begin{equation}\label{eqn:posterior_xi}
\xi|z_{1:n},x_{1:n} \sim \mathcal N(\mu_\xi, \Sigma_\xi),
\end{equation}
where the posterior mean $\mu_\xi$ and covariance matrix $\Sigma_\xi$ can be simply calculated according to \eqref{eqn:INNmodel1}. We leave the details of the derivation for the posterior distribution of $\xi$ in \ref{app:mu_sigma_xi}.  We have summarized in Algorithm 1 our NFF approach for learning and inferencing random fields.

\begin{remark}
Notice that if $k(x,\omega)$ is a scalar valued function with $D=1$, then $F_{KZ}(\cdot,x)$ is a one-dimensional invertible map and cannot be modeled by RealNVP. To fix this issue, we utilize NFF to model $\left(k(x,\omega),v(x,\omega)\right)^{\top}$ in experiments, where $v(x,\omega)  \stackrel{\mathrm{iid}}{\sim}  \mathcal{N}(0,1)$ is a Gaussian white noise (assumed to be unknown). Please see \ref{app:scalar_valued_nff} for more details.
\end{remark}

\begin{algorithm}[H]
\emph{\textbf{Learning:}}
\begin{itemize}
\item\textbf{1.} Specify the training set
\begin{equation*}
     \{k^s_{1:N}\}_{s=1}^{N_s} = \{k^s_{1},\ldots,k^s_{N}\}_{s=1}^{N_s} =\{(k(x_i^s,\omega^s))_{i=1}^{N}\}_{s=1}^{N_s}
\end{equation*}
\item\textbf{2.} Sample $N_b$ snapshots $\{(x^s_{1:N},k^s_{1:N})\}_{s=1}^{N_b}$ from the above training data
\item\textbf{3.} Calculate the loss $\mathcal{L_{\text{data}}}$ for $\{(x^s_{1:N},k^s_{1:N})\}_{s=1}^{N_b}$ via (\ref{eqn:INNmodelloss})
\item\textbf{4.} Let $W\leftarrow \text{Adam}(W-\eta\frac{\partial \mathcal{L_{\text{data}}}}{\partial W})$ to update all the involved parameters $W$ in (\ref{eqn:INNmodelloss}), $\eta$ is the learning rate
\item\textbf{5.} Repeat \textbf{Step 2-4} until convergence
\end{itemize}
\emph{\textbf{Inference/Prediction after learning:}}\\
\textbf{Input:} A snapshot of sensor measurements $\{k(x_1,\omega),\cdot\cdot\cdot,k(x_n,\omega)\}$\\
\textbf{Output:} Samples of the conditional distribution of $\{k(x_{n+1},\omega),\cdot\cdot\cdot,k(x_{n+n^{\prime}},\omega)\}$
\begin{itemize}
\item\textbf{1.} Calculate $z_i\triangleq z(x_i,\omega)=\mathcal{F}_{KZ}(k_i)$ for $i=1,\cdots,n$ given a snapshot of sensor measurements $\{k(x_1,\omega),\cdots,k(x_n,\omega)\}$
\item\textbf{2.} Calculate the posterior mean $\mu_{\xi}$ and covariance matrix matrix $\Sigma_{\xi}$ of $\xi$ according to (\ref{eqn:INNmodel1}) (see (\ref{eqn:mu_xi}, \ref{eqn:sigma_xi}) in \ref{app:mu_sigma_xi})
\item\textbf{3.} Sample $\xi\sim \mathcal{N}(\mu_{\xi},\Sigma_{\xi})$ and $\epsilon_i\sim \mathcal{N}(0,I)$ for $i=n+1,\cdots,n+n^{'}$
\item\textbf{4.} Calculate $k_i=k(x_i,\xi,\epsilon_i)$ for $i=n+1,\cdots,n+n^{'}$
\item\textbf{5.} Repeat \textbf{Steps 3-4}
\end{itemize}
\caption{Normalzing feild flow (NFF) for stochastic field}
\end{algorithm}

\subsection{Solving SDEs with physics informed normalizing field flow}
\label{S:3-3}
Next, we show how to use the NFF models to learning SDEs. To this end, we assume the measurements are given by (\ref{eqn:training_set}) and formalize the streamline of solving SDEs (\ref{eqn:SDE}) using normalizing field flow. For the general forward/inverse SDEs setting, three normalizing flows are first constructed to model the stochastic processes $k(x;\omega)$, $u(x;\omega)$ and $f(x;\omega)$, namely $\mathcal{F}_{ZK}(z_k,x)$, $\mathcal{F}_{ZU}(z_u,x))$ and $\mathcal{F}_{ZF}(z_f,x)$ based on Gaussian Processes $z_k(x,\xi,\epsilon_k), z_u(x,\xi,\epsilon_u)$ and $z_f(x,\xi,\epsilon_f)$ respectively. We denote those surrogates as $k(x,\xi,\epsilon_k), u(x,\xi,\epsilon_u)$ and $f(x,\xi,\epsilon_f)$.

\begin{figure}[htbp]
  \centering
  \includegraphics[width=0.9\linewidth]{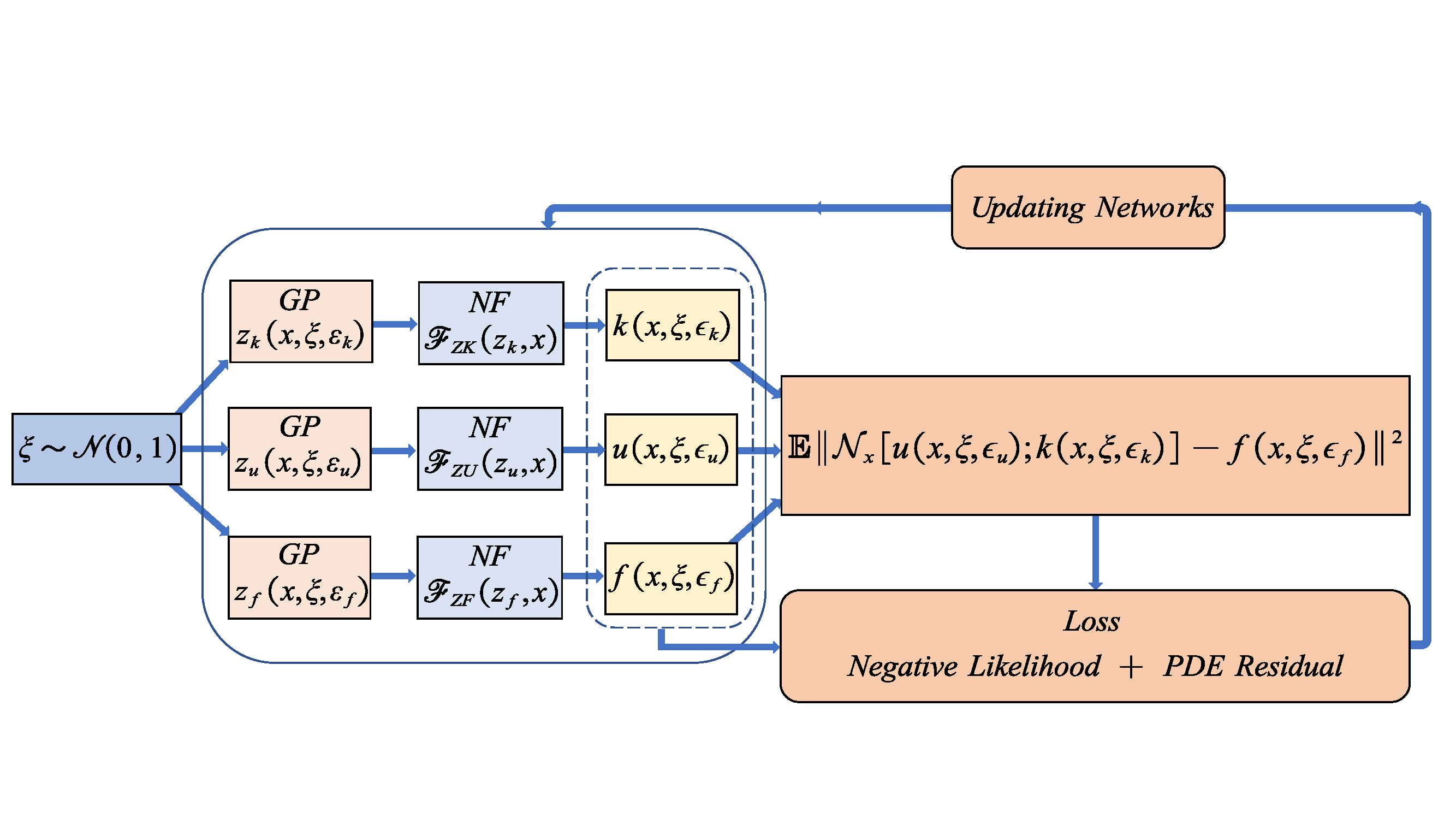}
  \caption{Schematic of physics informed Normalizing field flows for solving forward/inverse stochastic differential equations, where the loss function is given in Eq.~\ref{eqn:sdeloss}.}\label{fig:sketch}
\end{figure}

A sketch of the normalizing field flow for solving stochastic PDEs is given in Fig. \ref{fig:sketch}. Unlike the setting for learning a random field, here the loss function will be a weighted summation of three components: likelihood of the data, the weak formulation of SDE and boundary conditions. Namely, we shall include the physical loss (the SDEs) -- yielding the physics informed Normalizing field flow. The explicit expression for each of these three parts of loss function are given as following
\begin{equation}\label{eqn:sdeloss1}
\begin{array}{lll}
\mathcal{L}_{data}& = & -\frac{1}{N_{s}}\sum_{s=1}^{N_{s}}\log\mathbb{P}\bigg(k_{1:N_{k}}^{s},f_{1:N_{f}}^{s},u_{1:N_{u}}^{s}|x_{1:N_{k}}^{s},x_{1:N_{f}}^{s},x_{1:N_{u}}^{s}\bigg),\\
\mathcal{L}_{equ}& = & \mathbb{E}\Big\| \mathcal{N}_x[u(x,\xi,\epsilon_u);k(x,\xi,\epsilon_k)]-f(x,\xi,\epsilon_f)\big\|^2, \quad x\sim \mathcal{U}_{\mathcal{D}},\xi, \epsilon_u,\epsilon_k,\epsilon_f\sim \mathcal{N}(0,I),\\
\mathcal{L}_{bnd}& = & \mathbb{E}\big\| \mathcal{B}_x[u(x,\xi,\epsilon_u)]\big\|^2,\quad x\sim \mathcal{U}_{\Gamma}, \xi, \epsilon_u \sim\mathcal{N}(0,1).
\end{array}
\end{equation}
We then optimize the neural network parameters by minimizing the following total loss function:
\begin{equation}\label{eqn:sdeloss}
\mathcal{LOSS}=w_{data}\mathcal{L}_{data}+w_{equ}\mathcal{L}_{equ}+w_{bnd}\mathcal{L}_{bnd}.
\end{equation}
The proposed algorithm is summarized in Algorithm 2 and the schematic is plotted in Fig \ref{fig:sketch}. For the implementation, the equation loss $\mathcal{L}_{equ}$ is defined in the variational (weak) formulation (\cite{Kharazmi2020})
\begin{equation}\label{eqn:varsde}
(\mathcal{N}_x[u(x;\omega);k(x;\omega)]-f(x;\omega),h)=0, \forall h\in \text{test function set}.
\end{equation}
The details of the derivation of the associated loss functions are given in \ref{app:PI-NFF}.
\begin{algorithm}[H]
\textbf{1.} Specify the training set:
  \begin{equation*}
     \mathcal{S}_t =\{\{(x_{k,i}^{s},k_{i}^{s})\}_{i=1}^{N_{k}},\{(x_{u,i}^{s},u_{i}^{s})\}_{i=1}^{N_{u}},\{(x_{f,i}^{s},f_{i}^{s})\}_{i=1}^{N_{f}}\}_{s=1}^{N_{s}}.
  \end{equation*}
\\
\textbf{2.} Sample $N_b$ snapshots $\{(k_{1:N_k}^s,f_{1:N_f}^s,u_{1:N_u}^s)\}_{s=1}^{N_b}$ from the above training data. Following Algorithm 1 to built
three normalizing flows surrogates $k(x,\xi,\epsilon_k),u(x,\xi,\epsilon_u)$ and $f(x,\xi,\epsilon_f)$ respectively\\
\textbf{3.} Calculate $\mathcal{LOSS}=w_{data}\mathcal{L}_{data}+w_{equ}\mathcal{L}_{equ}+w_{bnd}\mathcal{L}_{bnd}$, where $\mathcal{L}_{data}$, $\mathcal{L}_{equ}$ and $\mathcal{L}_{bnd}$
are given by (\ref{eqn:sdeloss1})\\
\textbf{4.} Let $W\leftarrow \text{Adam}(W-\eta\frac{\partial \mathcal{LOSS}}{\partial W})$ to update all the involved parameters $W$\\
\textbf{5.} Repeat \textbf{2-4} until convergence
\caption{Physics-Informed NFF for solving stochastic PDEs}
\end{algorithm}

\section{Numerical results}
\label{S:4}
In this section, we shall test our NFF methods with several benchmark test cases. We first present the performance of the methods to approximate Non-Gaussian and mixed Non-Gaussian processes. We shall also present inference examples to illustrate the advantage of our flow-based generative model over the PI-GAN methods proposed in \cite{yangliusisc2020}. To further demonstrate the efficiency of the NFF strategy, we solve forward and inverse stochastic partial differential equations.  We use \texttt{ReLU} as the activation function in neural networks, since we use the variational form of the stochastic PDE and we do not need to take the higher order derivative compared with the strong formulation loss function. We assume the sensors are placed equidistantly in the physical domain for simplicity but the sensors can be random located for different snapshot during the training process, which is diffrent from PI-GAN. In our models, each NF consists of 6 transformation blocks with 128 neurons, while the DNNs used for the expansion coefficients have 4 layers with 128 neurons. The algorithms are implemented with the Adam optimizer in Pytorch where the learning rate is taken as 0.001.
\subsection{Application to approximate stochastic field}
\label{S:4-1}
We first present the performance of the NFF model for approximating Non-Gaussian stochastic processes and mixed Non-Gaussian processes.
\subsubsection{Case 1: Non-Gaussian stochastic field.}
\label{S:4-1-1}
Consider the following Non-Gaussian process
\begin{eqnarray}\label{eqn:Ngp}
k(x,\omega)=\text{exp}(\tilde{k}(x,\omega)),\quad \tilde{k}(x,\omega)\sim\mathcal{GP}\left(0, \sigma_c^2\exp\left(-\frac{(x - x')^2}{2l_c^2}\right)\right),\quad x,x'\in \mathcal{D}=[-1,1],
\end{eqnarray}
where $\sigma_c=1/\sqrt{2}$ is the standard deviation of $\tilde k$ and $l_c$ is the correlation length.
There are totally $12$ evenly spaced sensors. Moreover, for each snapshot, only measurements of randomly selecting $N_k$ ($N_k<12$) sensors are available.

\begin{figure}[htbp]
    \centering
    \includegraphics[width=0.9\linewidth]{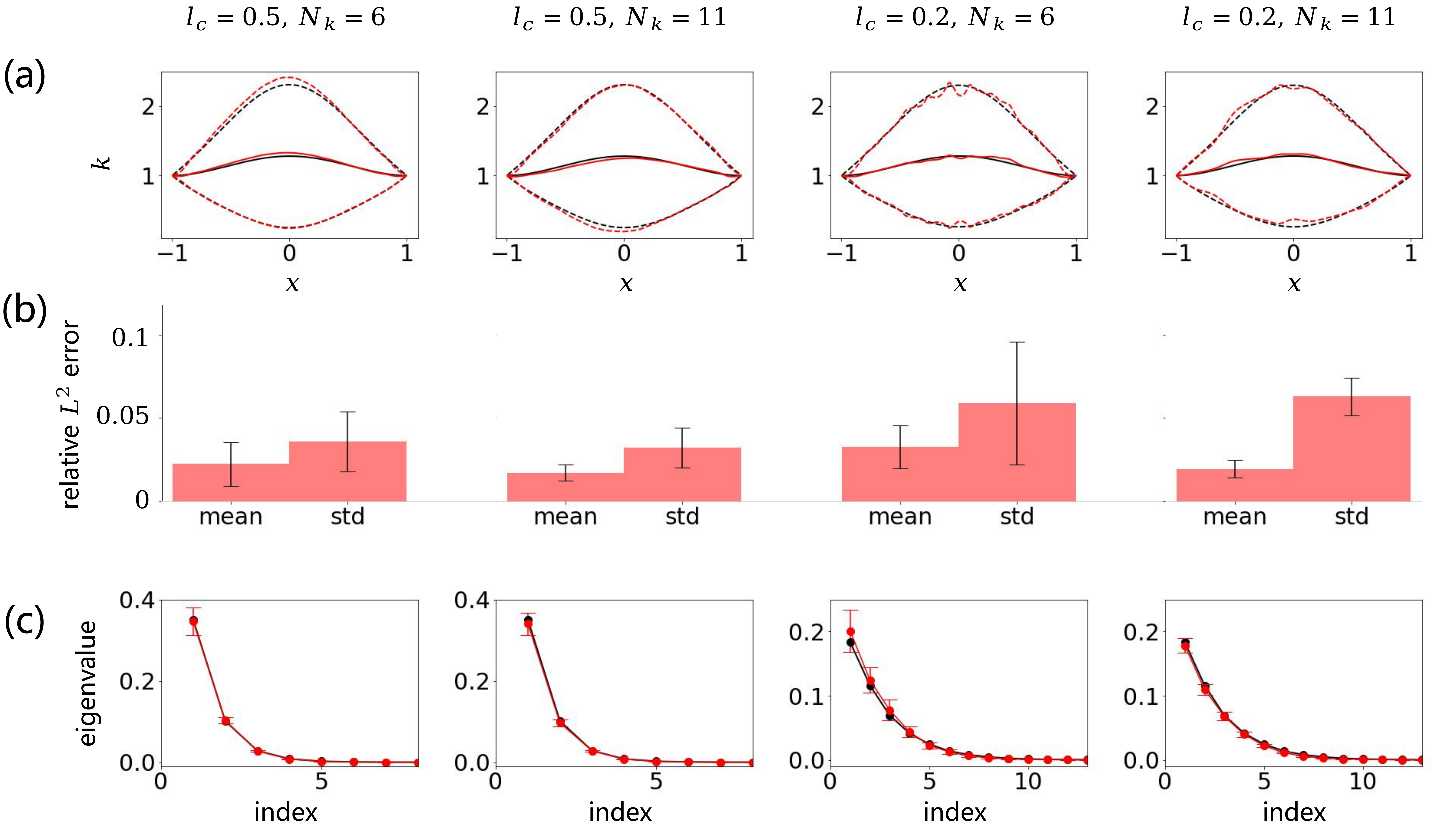}
    \caption{Estimation results of the Non-Gaussian stochastic field
    \eqref{eqn:Ngp} for different choices of the correlation length $l_c$ and the number of effective sensors $N_k$.
    (a) The mean $\mu(x)$ (solid lines) and standard deviation $\sigma(x)$ (dashed lines) of $k(x,\omega)$
    calculated from true models (black) and NFFs (red).
    (b) Relative errors of $\mu(x)$ and $\sigma(x)$ given by NFFs,
    where bars represent the averages and error bars over $10$ independent experiments.
    (c)
    Spectra of the correlation structure for the generated processes of different correlation, where the black lines represent the true values of the top eigenvalues, and the red lines indicate the corresponding
    mean values and standard errors of the estimated eigenvalues calculated from $10$ independent experiments.}\label{fig:approximation_NGP}
\end{figure}

We consider here the choices $l_c=0.5,0.2$ and $N_k=6,11$.
For each choice,
the number of snapshots is $10^3$,
and we stop the training after 400 epochs with batch size $128$. The number of KL expansion is taken as $M=30$.
We generate $1\times 10^5$ sample paths from the trained NFF model and calculate its spectra, i.e., the eigenvalues of the covariance matrices from the principal component analysis. The results are summarized in Fig.~\ref{fig:approximation_NGP}, where the relative errors of estimated mean and standard deviation of $k(x,\omega)$ are calculated as
\begin{eqnarray}\label{eqn:error}
\text{relative error in mean} & = & \frac{\left\Vert \mu(x)-\hat{\mu}(x)\right\Vert _{2}}{\left\Vert \mu(x)\right\Vert _{2}},\\
\text{relative error in std} & = & \frac{\left\Vert \sigma(x)-\hat{\sigma}(x)\right\Vert _{2}}{\left\Vert \sigma(x)\right\Vert _{2}},
\end{eqnarray}
with $\mu(x)=\mathbb E_\omega[k(x,\omega)]$ and
$\sigma(x)=\mathrm{cov}_\omega[k(x,\omega)]$. It is noticed that the learned field matches well with the target field, even with very few sensor locations.

\subsubsection{Case 2: Mixed Non-Gaussian stochastic field}
\label{S:4-1-2}
\begin{figure}[htbp]
    \centering
    \includegraphics[width=0.75\linewidth]{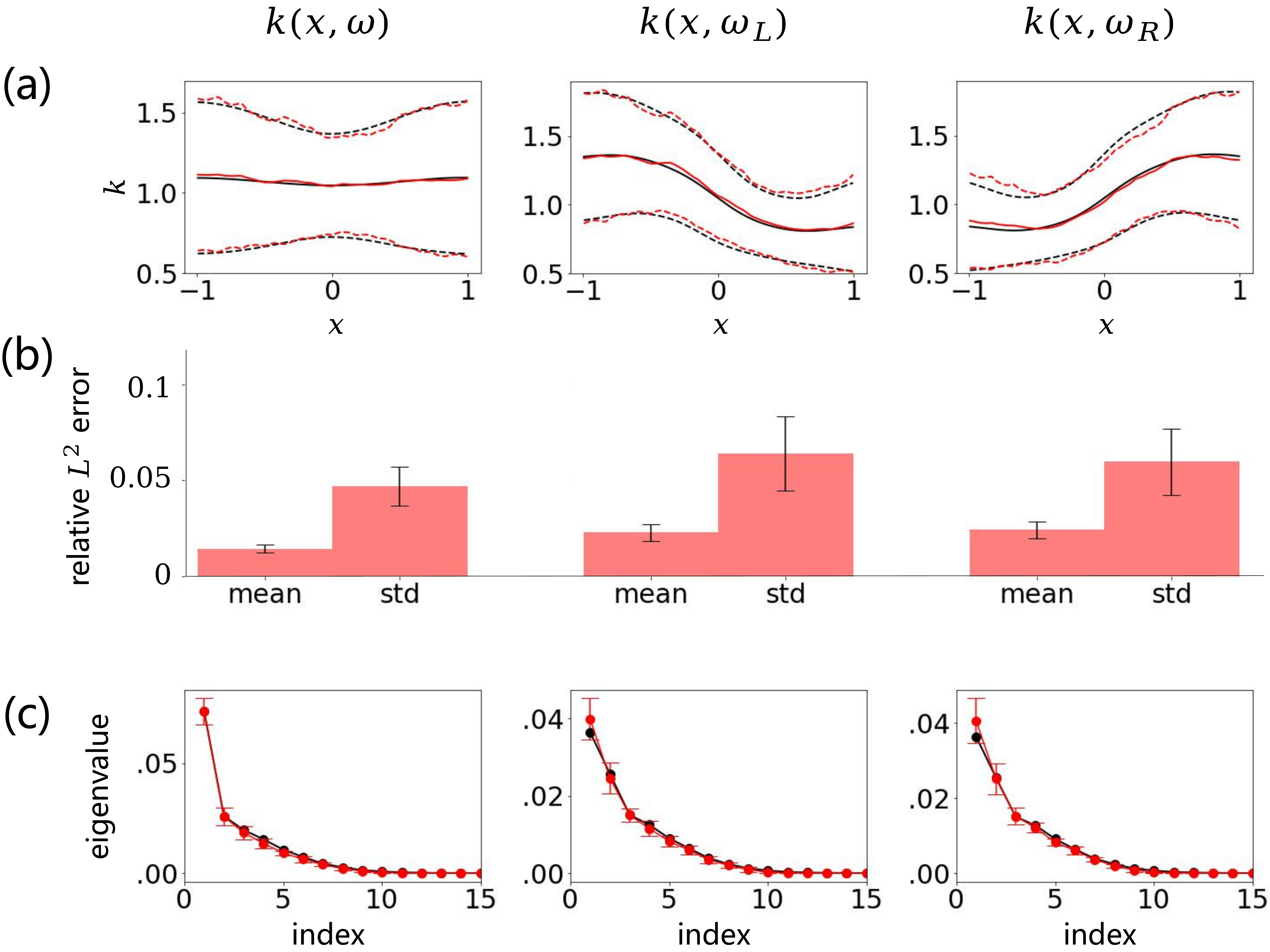}
    \caption{Estimation results of the Mixed Non-Gaussian stochastic field
    \eqref{eqn:Mgp}.
    (a) The mean $\mu(x)$ (solid lines) and standard deviation $\sigma(x)$ (dashed lines)
    calculated from true models (black) and NFFs (red).
    (b) Relative errors of $\mu(x)$ and $\sigma(x)$ given by NFFs,
    where bars represent the averages and error bars over $10$ independent experiments.
    (c)
    Spectra of the correlation structure for the generated processes, where the black lines represent the true values of the top eigenvalues, and the red lines indicate the corresponding
    mean values and standard errors of the estimated eigenvalues calculated from $10$ independent experiments.}\label{fig:approximation_MGP}
\end{figure}

To further demonstrate the efficiency of the NFF model, we consider the approximation of the following mixed Non-Gaussian stochastic field with
\begin{equation}\label{eqn:Mgp}
k(x,\omega)=\exp\left(\frac{3}{10}\left(\tilde{k}(x,\omega)+m(x,\omega)\right)\right),
\end{equation}
where $\tilde{k}(x,\omega)$ is defined as in (\ref{eqn:Ngp}) with $\sigma_c=1$ and $l_c=0.2$, and
\begin{equation}\label{eqn:Mgp1}
m(x,\omega)=\left\{ \begin{array}{ll}
\sin\frac{\pi x}{2}, & \text{with prob. }0.5,\\
-\sin\frac{\pi x}{2}, & \text{with prob. }0.5.
\end{array}\right.
\end{equation}
We generate $10^3$ snapshots with $13$ sensors equally spaced in $[-1,1]$, where only $N_k=7$ sensor measurements are collected for each snapshot.
The batch size in all our test cases is $128$. The Number of KL expansion is taken as $M=30$. We stop the training after $400$ epochs. The results are illustrated in the first column of Fig.~\ref{fig:approximation_MGP}.

As can be seen from \eqref{eqn:Mgp}, $k(x,\omega)$ has two modes
\begin{eqnarray}
\omega_L & \sim & \omega\left\vert\int_{-1}^{0}k(x,\omega)\mathrm{d}x\ge\int_{0}^{1}k(x,\omega)\mathrm{d}x\right.,\label{eq:omega_L}\\
\omega_R & \sim & \omega\left\vert\int_{-1}^{0}k(x,\omega)\mathrm{d}x<\int_{0}^{1}k(x,\omega)\mathrm{d}x\right..\label{eq:omega_R}
\end{eqnarray}
Hence, we also compare the statistical properties of $k(x,\omega)$ conditional on the two modes and those estimated by the NFF in the second and third columns of Fig.~\ref{fig:approximation_MGP}. Again, the learned field yields good agreements with the target field.

\subsubsection{Case 3: Inference and prediction after learning}
\label{S:4-1-3}
\begin{figure}[htbp]
    \centering
    \includegraphics[width=0.75\linewidth]{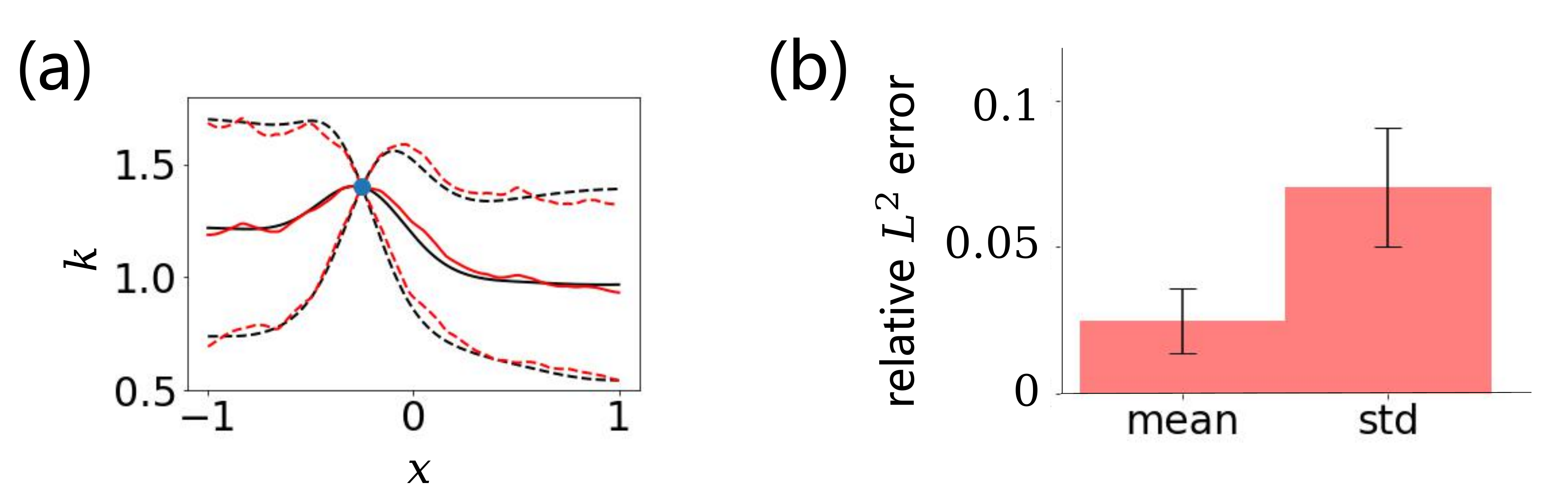}
    \caption{(a) The posterior mean $\mu(x)=\mathbb E[k(x,\omega)|k(x_1,\omega)=k_1]$ (solid lines) and standard deviation $\sigma(x)=\mathrm{cov}[k(x,\omega)|k(x_1,\omega)=k_1]$ (dashed lines) of $k(x,\omega)$ for given $k(x_1,\omega)=k_1$ calculated from true models (black) and the inference results via the NFF (red). (b) Relative errors of $\mu(x)$ and $\sigma(x)$ given by NFFs,
    where bars represent the averages and error bars over $10$ independent experiments.}\label{fig:inference_MGP}
\end{figure}

In this section, we show the inference/prediction capability of the NFF model. According to Algorithm 1, given a snapshot of $n$ measurements $\{k(x_1,\omega),\cdot\cdot\cdot,k(x_n,\omega)\}$, we can compute the conditional distributions of new $n^{\prime}$ locations after leaning. We assume we have a random measurement located at $x\in[-1,1]$ from mixed Non-Gaussian process in (\ref{eqn:Mgp}), then we infer the posterior mean and variance on new location. We plot the inferred mean and variance in Fig.\ref{fig:inference_MGP} from the inference models, which show good match with the true value. This indicates that the NFF models admit very good capability for inference/prediction.

\subsection{Application to 1-d SDEs}
\label{S:4-2}
We consider the following one-dimensional stochastic elliptic equation:
\begin{equation}\label{eqn:stochastic_elliptic_1d}
\begin{gathered}
    -\dv{x}\left(k(x;\omega)\dv{x}u(x;\omega)\right) = f(x;\omega), \quad x\in[-1, 1] \text{ and } \omega \in \Omega,\\
    u(-1) = u(1) = 0.
\end{gathered}
\end{equation}
The randomness comes from the mixed Non-Gaussian $k(x;\omega)$ and the Gaussian forcing term $f(x;\omega)$, which are modeled as the following stochastic processes:
\begin{equation}\label{eqn:sde1d_kf}
\begin{array}{ccc}
k(x,\omega)& = & \text{exp}\bigg [\frac{3}{10}(\tilde{k}(x,\omega)+m(x,\omega))\bigg ],\\
f(x,\omega)& \sim & \mathcal{GP}\bigg(\frac{1}{2}, \frac{9}{400}\exp\left(-25(x - x')^2\right)\bigg),
\end{array}
\end{equation}
where $\tilde{k}(x,\omega)$ is defined as in (\ref{eqn:Ngp}) with $\sigma_c=1$ and $l_c=0.2$, and $m(x,\omega)$ is given by (\ref{eqn:Mgp1}). We will demonstrate the effectiveness of solving (\ref{eqn:stochastic_elliptic_1d}) with the
physics-informed NFF Algorithm 2 for both forward and inverse problems.

\subsubsection{Case 1: Forward problem.}
\label{S:4-2-1}
\begin{figure}[htbp]
    \centering
    \includegraphics[width=0.75\linewidth]{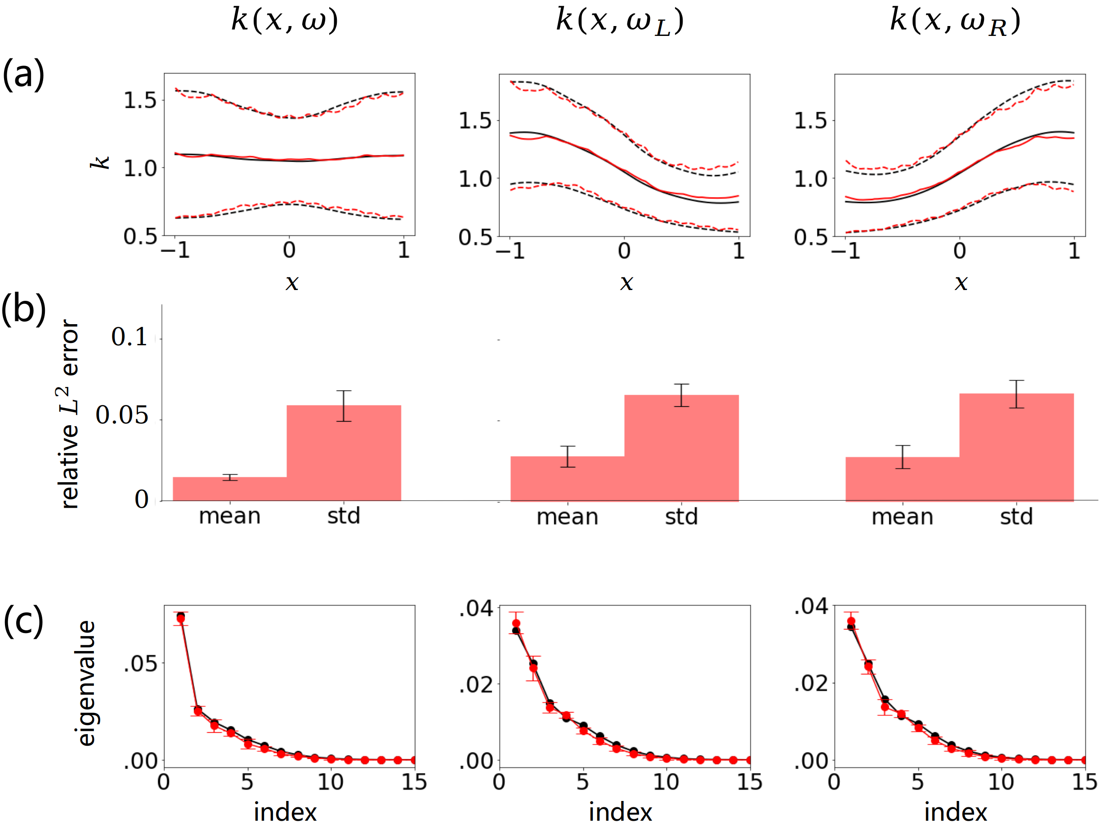}
    \caption{Computation results of $k(x,\omega)$ for the forward stochastic PDE
    \eqref{eqn:stochastic_elliptic_1d}.
    (a) The mean $\mu(x)$ (solid lines) and standard deviation $\sigma(x)$ (dashed lines)
    calculated from MC simulation (black) and NFFs (red).
    (b) Relative errors of $\mu(x)$ and $\sigma(x)$ given by NFFs,
    where bars represent the averages and error bars over $10$ independent experiments.
    (c)
    Spectra of the correlation structure for the generated processes, where the black lines represent the values of the top eigenvalues calculated via the MC simulation, and the red lines indicate the corresponding
    mean values and standard errors of the estimated eigenvalues calculated from $10$ independent experiments.}\label{fig:forward_problem_k}
\end{figure}

\begin{figure}[htbp]
    \centering
    \includegraphics[width=0.75\linewidth]{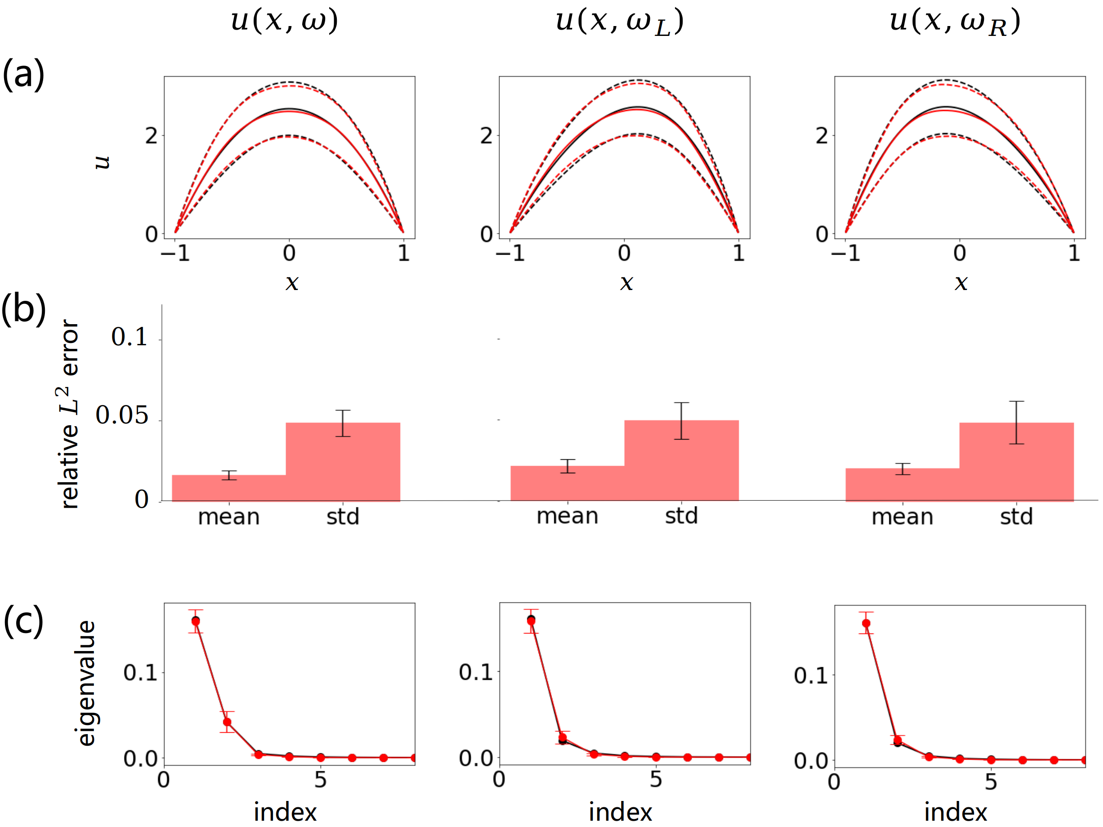}
    \caption{Computation results of $u(x,\omega)$ for the forward stochastic PDE
    \eqref{eqn:stochastic_elliptic_1d}.
    (a) The mean $\mu(x)$ (solid lines) and standard deviation $\sigma(x)$ (dashed lines)
    calculated from MC simulation (black) and NFFs (red).
    (b) Relative errors of $\mu(x)$ and $\sigma(x)$ given by NFFs,
    where bars represent the averages and error bars over $10$ independent experiments.
    (c)
    Spectra of the correlation structure for the generated processes, where the black lines represent the true values of the top eigenvalues calculated via the MC simulation, and the red lines indicate the corresponding
    mean values and standard errors of the estimated eigenvalues calculated from $10$ independent experiments.}\label{fig:forward_problem_u}
\end{figure}

In this case, we place $13$  sensors of $k(x,\omega)$ and $21$ sensors of $f(x,\omega)$ in the physical domain. For each snapshot, measurements of $N_k=7$ sensors of $k(x,\omega)$ and $N_k=11$ sensors of $f(x,\omega)$ are available. The results are shown in Fig.~\ref{fig:forward_problem_k} and Fig.~\ref{fig:forward_problem_u} respectively, including the mean and standard variation calculated via the physics informed NFF model, the relative $L^2$ error and the spectra of the correlation structure for the generated process.


\subsubsection{Case 2: Inverse and mixed problems.}
\label{S:4-2-2}
\begin{figure}[htbp]
    \centering
    \includegraphics[width=\linewidth]{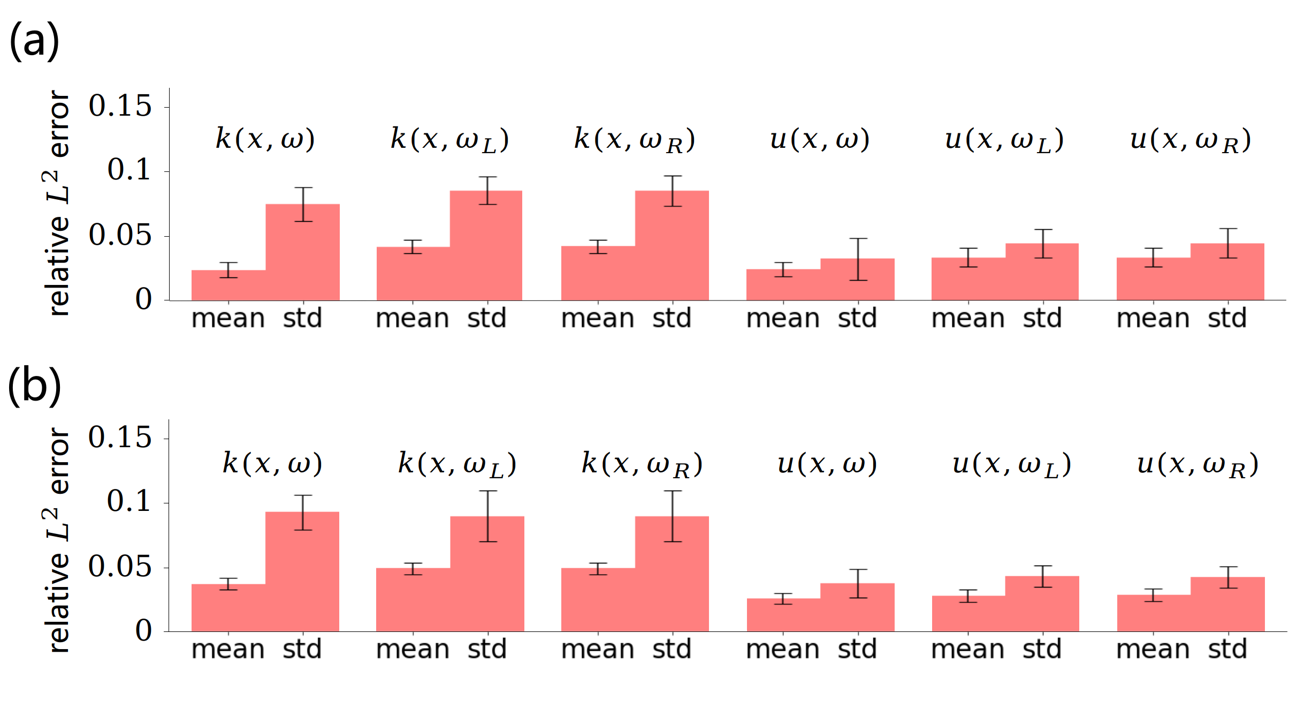}
    \caption{Relative errors of the inferred mean and standard deviation for both $k(x,\omega)$ and $u(x,\omega)$. (a) inverse problem and (b) mixed problem, where bars represent the averages and error bars over $10$ independent experiments.}\label{fig:approximation_inverse_hybrid}
\end{figure}

We solve (\ref{eqn:stochastic_elliptic_1d}) here again but now we assume that we have some extra information on the solution $u(x;\omega)$ but incomplete information of the
diffusion coefficient $k(x;\omega)$. In addition, $f(x,\omega)\equiv 1$. Specifically, we consider two scenarios of sensor placements: 1): inverse problem: $N_k=1$ (at $x=0$), $N_u=7$ (selected from $13$ sensors); 2): mixed problem: $N_k=3$ (selected from $5$ sensors), $N_u=5$ (selected from $9$ sensors). we use 1000 snapshots for training and set the input random dimension $M$ to be 40. In Fig.~\ref{fig:approximation_inverse_hybrid}, we compare the relative errors with reference solutions calculated
from Monte Carlo sample paths.

For both cases, we can see that the numerical errors are in the same order of magnitude with the MC method, showing the effectiveness of the NFF models for solving forward/inverse SDE problems.

\subsection{Application to 2-d SDEs}
\begin{figure}[htbp]
    \centering
    \includegraphics[width=0.75\linewidth]{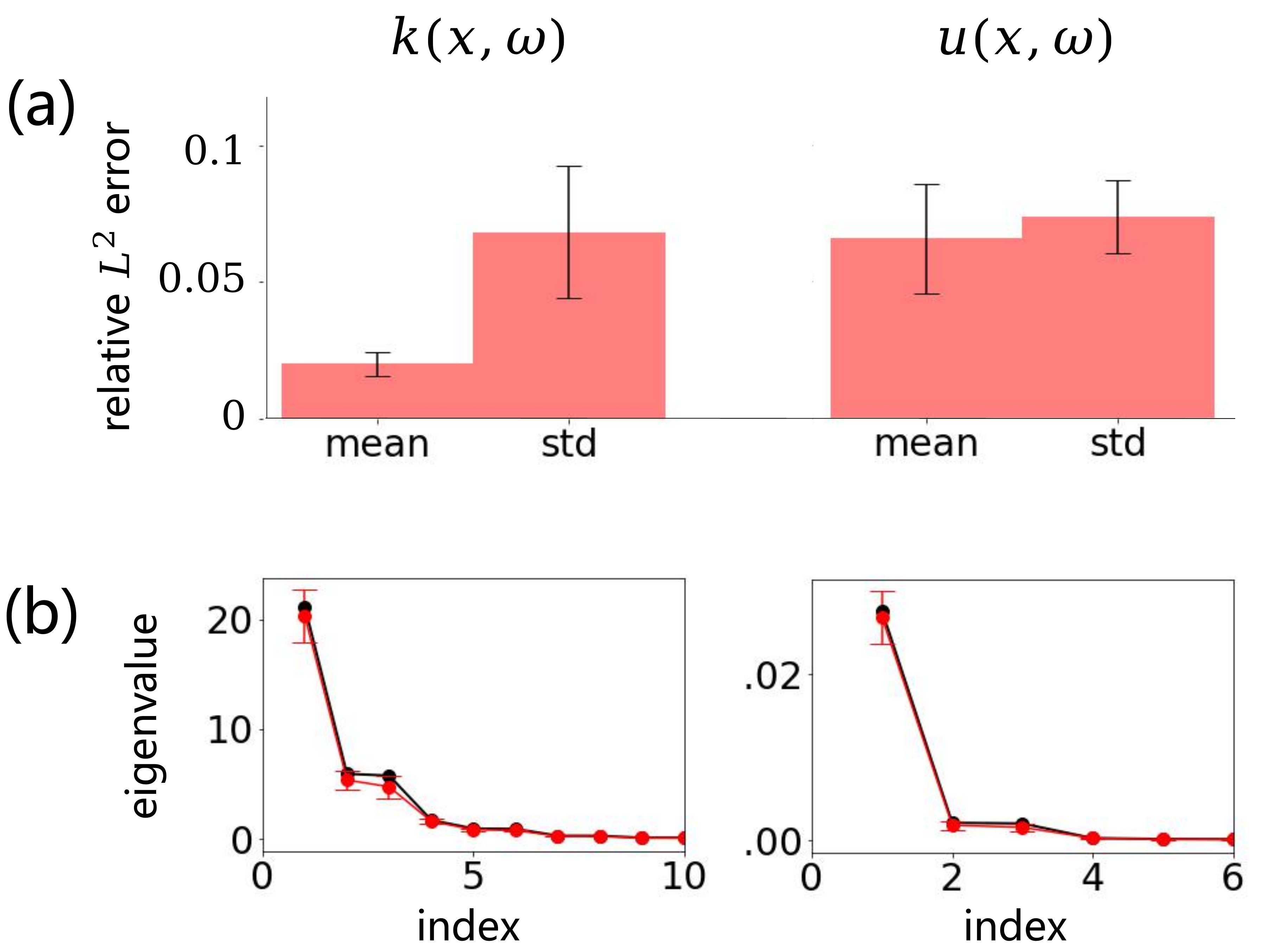}
    \caption{(a) Relative errors of means and standard deviations of $k(x,\omega), u(x,\omega)$  given by NFFs,
    where bars represent the averages and error bars over $10$ independent experiments.
    (b)
    Spectra of the correlation structure for the generated processes, where the black lines represent the values of the top eigenvalues calculated via the MC simulation, and the red lines indicate the corresponding
    mean values and standard errors of the estimated eigenvalues calculated from $10$ independent experiments.}
    \label{fig:two_dimensional_forward}
\end{figure}

We finally consider the two-dimensional stochastic elliptic equation as following
$$-\nabla\cdot(k(x,\omega)\nabla u(x,\omega))=1,\quad x\in \mathcal D=[0,1]^2,$$
$$ u(x,\omega)=0,\quad x\in \partial \mathcal D,$$
where $k(x,\omega) = \exp\left(\tilde{k} (x,\omega)\right)$ and
\[\tilde{k}(x,\omega)\sim \mathcal{GP}\left(0, \exp(-16\Vert x_1 - x_2\Vert^2)\right).\]
we place $51$ sensors of $k(x,\omega)$ in the physical domain,
and generate $1000$ snapshots for training, where
measurements of $N_k=30$ sensors
are available for each snapshot. The results of applying the physics informed NFF to the forward problem are shown in Fig.~\ref{fig:two_dimensional_forward}.

\section{Summary}
\label{S:5}

We have presented the \textit{normalizing field flows} for learning random fields from scattered measurements. Our NFF model is constructed by using a bijective transformation  between a tractable Gaussian random field and the target stochastic field. The NFF model is fully data driven, and can be used to solve data-driven forward, inverse, and mixed stochastic partial differential equations in a unified framework. Moreover, unlike the setting of the GAN based solvers for SDEs \cite{yangliusisc2020}, sensor locations are not assumed to be fixed for different snapshots in our flow model. When the physics informed NFF model is adopted to learn random fields in UQ problems, it can alleviate the \textit{curse of dimensionality} that appears in  most traditional approaches such as polynomial chaos. In our future works, we shall use the NFF model to deal with time-dependent problems and problems with complex solution structures (such as multi-scales).

\section*{Acknowledgement}

The first author is supported by the NSF of China (under grant numbers 12071301, 11671265) and the Shanghai Municipal Science and Technology Commission (No.20JC1412500). The second author is supported by the NSF of China (under grant number 12171367), the Shanghai Municipal Science and Technology Commission (No.20JC1413500) and the fundamental research funds for the central universities of China (No.22120210133). The last author is supported by the National Key R\&D Program of China (2020YFA0712000), the NSF of China (under grant numbers 11822111, 11688101), the science challenge project (No.TZ2018001), and youth innovation promotion association (CAS).

\FloatBarrier

\section*{References}
\bibliography{reference_NNF}

\newpage
\appendix

\section{Calculation of $\boldsymbol{\mu}_\xi$ and $\boldsymbol{\Sigma}_\xi$}\label{app:mu_sigma_xi}
Let $\tilde z_i=z_i-A(x_i)$. According to \eqref{eqn:INNmodel1}, the joint distribution of $\xi$ and $\tilde z_{1:n}$ is a multivariate normal distribution with mean zero and covariance matrix
\begin{equation*}
\mathrm{cov}\left(\begin{array}{c}
\xi\\
\tilde{z}_{1}\\
\vdots\\
\tilde{z}_{N}
\end{array}\right)=\left[\begin{array}{cc}
\mathbf{I}_{M} & B(\mathbf{X})^{\top}\\
B(\mathbf{X}) & B(\mathbf{X})B(\mathbf{X})^{\top}+\mathrm{diag}\left(C(\mathbf{X})\right)^{2}
\end{array}\right],
\end{equation*}where
\begin{equation*}
B(\mathbf{X})=\left(\begin{array}{c}
B(x_{1})\\
\vdots\\
B(x_{n})
\end{array}\right),\quad C(\mathbf{X})=\left(\begin{array}{c}
C(x_{1})\\
\vdots\\
C(x_{n})
\end{array}\right).
\end{equation*}
Thus,
\begin{eqnarray}
\boldsymbol{\mu}_{\xi} & = & B(\mathbf{X})^{\top}\left(B(\mathbf{X})B(\mathbf{X})^{\top}+\mathrm{diag}\left(C(\mathbf{X})\right)^{2}\right)^{-1}\tilde{\mathbf{Z}},\label{eqn:mu_xi}\\
\boldsymbol{\Sigma}_{\xi} & = & \mathbf{I}_{M}-B(\mathbf{X})^{\top}\left(B(\mathbf{X})B(\mathbf{X})^{\top}+\mathrm{diag}\left(C(\mathbf{X})\right)^{2}\right)^{-1}B(\mathbf{X})\label{eqn:sigma_xi}
\end{eqnarray}
with
\[
\tilde{\mathbf{Z}}=\left(\begin{array}{c}
z_{1}-A(x_{1})\\
\vdots\\
z_{n}-A(x_{n})
\end{array}\right).
\]
\section{Likelihood for scalar valued NFF}\label{app:scalar_valued_nff}
For $D=1$, we model $\left(k(x,\omega),v(x,\omega)\right)$ as
\[\left(k(x,\omega),v(x,\omega)\right)=\mathcal{F}_{ZK}\left(z(x,\omega),\zeta(x,\omega),x\right),\]where $z(x,\omega)$ is a Gaussian process as defined in \eqref{eqn:INNmodel1}, $\zeta(x,\omega)\stackrel{\mathrm{iid}}{\sim}\mathcal N(0,1)$, and $\mathcal{F}_{ZK}\left(\cdot,\cdot,x\right)=\mathcal{F}_{KZ}^{-1}\left(\cdot,\cdot,x\right)$ is an invertible map from $\mathbb R^2$ to $\mathbb R^2$ constructed by a RealNVP.

For a snapshot of sensor measurements $k_{1:N}^s$, we can draw $v_{1:N}^s$ independently from $\mathcal N(0,1)$ and calculate the log-likelihood of $k_{1:N}^s$ and $v_{1:N}^s$ as
\begin{equation}\label{eqn:ll_kv}
\log\mathbb{P}(k_{1:N}^{s},v_{1:N}^{s}|x_{1:N}^{s})=\log\mathcal{N}(z_{1:N}^{s}|\boldsymbol{\mu}^{s},\boldsymbol{\mathbf{\Sigma}}^{s})+\sum_{i=1}^{N}\log\mathcal{N}(\zeta_{i}|0,1)+\sum_{i=1}^{N}\log\left|\mathrm{det}\frac{\partial \mathcal{F}_{KZ}(k_{i}^{s},v_{i}^{s},x_{i}^{s})}{\partial(k_{i}^{s},v_{i}^{s})}\right|.
\end{equation}
The corresponding NFF method is summarized in Algorithm 3.
\begin{algorithm}[H]
\emph{\textbf{Learning:}}
\begin{itemize}
\item\textbf{1.} Specify the training set
\begin{equation*}
     \{k^s_{1:N}\}_{s=1}^{N_s} = \{(k^s_{1},\ldots,k^s_{N}\}_{s=1}^{N_s} =\{(k(x_i^s,\omega^s))_{i=1}^{N}\}_{s=1}^{N_s}
\end{equation*}
\item\textbf{2.} Sample $N_b$ snapshots $\{(x^s_{1:N},k^s_{1:N})\}_{s=1}^{N_b}$ from the above training data, and sample $\{(v^s_{1:N})\}_{s=1}^{N_b}$ from $\mathcal N(0,1)$
\item\textbf{3.} Calculate the loss \[\mathcal{L_{\text{data}}}=-\frac{1}{N_{b}}\sum_{s=1}^{N_{b}}\log\mathbb{P}(k_{1:N}^{s},v_{1:N}^{s}|x_{1:N}^{s})\] for $\{(x^s_{1:N},k^s_{1:N},v^s_{1:N})\}_{s=1}^{N_b}$ via (\ref{eqn:ll_kv})
\item\textbf{4.} Let $W\leftarrow \text{Adam}(W-\eta\frac{\partial \mathcal{L_{\text{data}}}}{\partial W})$ to update all the involved parameters $W$ in (\ref{eqn:INNmodelloss}), $\eta$ is the learning rate
\item\textbf{5.} Repeat \textbf{Step 2-4} until convergence
\end{itemize}
\emph{\textbf{Inference/Prediction after learning:}}\\
\textbf{Input:} A snapshot of sensor measurements $\{k(x_1,\omega),\cdot\cdot\cdot,k(x_n,\omega)\}$\\
\textbf{Output:} Samples of the conditional distribution of $\{k(x_{n+1},\omega),\cdot\cdot\cdot,k(x_{n+n^{\prime}},\omega)\}$
\begin{itemize}
\item\textbf{1.} Draw $v_i\sim\mathcal N(0,1)$ and calculate $(z_i,\zeta_i)=\mathcal{F}_{KZ}(k_i,v_i,x)$ for $i=1,\cdots,n$ given a snapshot of sensor measurements $\{k(x_1,\omega),\cdots,k(x_n,\omega)\}$
\item\textbf{2.} Calculate the posterior mean $\mu_{\xi}$ and covariance matrix matrix $\Sigma_{\xi}$ of $\xi$ by (\ref{eqn:mu_xi}, \ref{eqn:sigma_xi})
\item\textbf{3.} Sample $\xi\sim \mathcal{N}(\mu_{\xi},\Sigma_{\xi})$ and $\epsilon_i\sim \mathcal{N}(0,I)$ for $i=n+1,\cdots,n+n^{'}$
\item\textbf{4.} Calculate $z_i=z(x_i,\xi,\epsilon_i)$ for $i=n+1,\cdots,n+n^{'}$
\item\textbf{5.} Sample $\zeta_i\sim \mathcal{N}(0,I)$ and calculate $(k_i,v_i)=\mathcal{F}_{ZK}(z_i,\zeta_i,x_i)$ for $i=n+1,\cdots,n+n^{'}$
\item\textbf{6.} Repeat \textbf{Steps 1-5}
\end{itemize}
\caption{NFF for scalar valued stochastic field}
\end{algorithm}

\section{Loss functions for physics-informed NFF}\label{app:PI-NFF}
\subsection{Likelihood loss}
For a snapshot $s$, we have
\((z_{k,1:N_{k}}^{s},z_{f,1:N_{f}}^{s},z_{u,1:N_{u}}^{s})\sim\mathcal{N}(\boldsymbol{\mu}^{s},\boldsymbol{\Sigma}^{s})\)
with
\[\boldsymbol{\mu}^{s}=\left(\begin{array}{c}
A(x_{k,1}^{s})\\
\vdots\\
A(x_{k,N_{k}}^{s})\\
A(x_{f,1}^{s})\\
\vdots\\
A(x_{f,N_{f}}^{s})\\
A(x_{u,1}^{s})\\
\vdots\\
A(x_{u,N_{u}}^{s})
\end{array}\right),\quad\boldsymbol{\Sigma}^{s}=\left(\begin{array}{c}
B(x_{k,1}^{s})\\
\vdots\\
B(x_{k,N_{k}}^{s})\\
B(x_{f,1}^{s})\\
\vdots\\
B(x_{f,N_{f}}^{s})\\
B(x_{u,1}^{s})\\
\vdots\\
B(x_{u,N_{u}}^{s})
\end{array}\right)\left(\begin{array}{c}
B(x_{k,1}^{s})\\
\vdots\\
B(x_{k,N_{k}}^{s})\\
B(x_{f,1}^{s})\\
\vdots\\
B(x_{f,N_{f}}^{s})\\
B(x_{u,1}^{s})\\
\vdots\\
B(x_{u,N_{u}}^{s})
\end{array}\right)^{\top}+\text{diag}\left(\begin{array}{c}
C(x_{k,1}^{s})\\
\vdots\\
C(x_{k,N_{k}}^{s})\\
C(x_{f,1}^{s})\\
\vdots\\
C(x_{f,N_{f}}^{s})\\
C(x_{u,1}^{s})\\
\vdots\\
C(x_{u,N_{u}}^{s})
\end{array}\right)^{2},\]
and
\begin{eqnarray*}
\log\mathbb{P}\bigg(k_{1:N_{k}}^{s},f_{1:N_{f}}^{s},u_{1:N_{u}}^{s}|x_{k,1:N_{k}}^{s},x_{f,1:N_{f}}^{s},x_{u,1:N_{u}}^{s}\bigg) & = & \log\mathcal{N}\bigg(z_{k,1:N_{k}}^{s},z_{f,1:N_{f}}^{s},z_{u,1:N_{u}}^{s}|\boldsymbol{\mu}^{s},\boldsymbol{\Sigma}^{s}\bigg)\\
 &  & +\sum_{i=1}^{N_{k}}\log\left|\mathrm{det}\frac{\partial \mathcal{F}_{KZ}(k_{i}^{s},x_{k,i}^{s})}{\partial k_{i}^{s}}\right|\\
 &  & +\sum_{i=1}^{N_{f}}\log\left|\mathrm{det}\frac{\partial \mathcal{F}_{FZ}(f_{i}^{s},x_{f,i}^{s})}{\partial f_{i}^{s}}\right|\\
 &  & +\sum_{i=1}^{N_{u}}\log\left|\mathrm{det}\frac{\partial \mathcal{F}_{UZ}(u_{i}^{s},x_{u,i}^{s})}{\partial u_{i}^{s}}\right|.
\end{eqnarray*}

\subsection{Likelihood for scalar valued NFFs}
Suppose that $k(x,\omega),f(x,\omega),u(x,\omega)$ are scalar valued, and $\mathcal{F}_{KZ},\mathcal{F}_{FZ},\mathcal{F}_{UZ}$ are all constructed by normalizing flows with two inputs and two outputs. For $N_b$ snapshots $\left\{k_{1:N_{k}}^{s},f_{1:N_{f}}^{s},u_{1:N_{u}}^{s}\right\}_{s=1}^{N_b}$ sampled from the training data, we can draw $\left\{v_{k,1:N_{k}}^{s},v_{f,1:N_{f}}^{s},v_{u,1:N_{u}}^{s}\right\}_{s=1}^{N_b}$ from $\mathcal N(0,1)$, let
\begin{eqnarray*}
(z_{k,i}^{s},\zeta_{k,i}^{s}) & = & \mathcal{F}_{KZ}(k_{i}^{s},v_{k,i}^{s}),\\
(z_{f,i}^{s},\zeta_{f,i}^{s}) & = & \mathcal F_{FZ}(f_{i}^{s},v_{f,i}^{s}),\\
(z_{u,i}^{s},\zeta_{u,i}^{s}) & = & \mathcal F_{UZ}(u_{i}^{s},v_{u,i}^{s}),
\end{eqnarray*}
and calculate
\[\mathcal{L}_{\mathrm{data}}=-\frac{1}{N_{b}}\sum_{s=1}^{N_{b}}\log\mathbb{P}\left(k_{1:N_{k}}^{s},v_{k,1:N_{k}}^{s},f_{1:N_{f}}^{s},v_{f,1:N_{f}}^{s},u_{1:N_{u}}^{s},v_{u,1:N_{u}}^{s}|x_{k,1:N_{k}}^{s},x_{f,1:N_{f}}^{s},x_{u,1:N_{u}}^{s}\right)\]
with
\begin{eqnarray*}
\log\mathbb{P}\bigg(k_{1:N_{k}}^{s},v_{k,1:N_{k}}^{s},f_{1:N_{f}}^{s},v_{f,1:N_{f}}^{s},u_{1:N_{u}}^{s},v_{u,1:N_{u}}^{s}|x_{k,1:N_{k}}^{s},x_{k,1:N_{f}}^{s},x_{k,1:N_{u}}^{s}\bigg) & = & \log\mathcal{N}\bigg(z_{k,1:N_{k}}^{s},z_{f,1:N_{f}}^{s},z_{u,1:N_{u}}^{s}|\boldsymbol{\mu}^{s},\boldsymbol{\Sigma}^{s}\bigg)\\
 &  & +\log\mathcal{N}\left(\zeta_{k,1:N_{k}}^{s},\zeta_{f,1:N_{f}}^{s},\zeta_{u,1:N_{u}}^{s}|\mathbf{0},\mathbf{I}_{N_{k}+N_{f}+N_{u}}\right)\\
 &  & +\sum_{i=1}^{N_{k}}\log\left|\mathrm{det}\frac{\partial \mathcal{F}_{KZ}(k_{i}^{s},x_{k,i}^{s})}{\partial k_{i}^{s}}\right|\\
 &  & +\sum_{i=1}^{N_{f}}\log\left|\mathrm{det}\frac{\partial \mathcal{F}_{FZ}(f_{i}^{s},x_{f,i}^{s})}{\partial f_{i}^{s}}\right|\\
 &  & +\sum_{i=1}^{N_{u}}\log\left|\mathrm{det}\frac{\partial \mathcal{F}_{UZ}(u_{i}^{s},x_{u,i}^{s})}{\partial u_{i}^{s}}\right|
\end{eqnarray*}

\subsection{Equation loss}
Here we only consider the stochastic elliptic equation
\[-\nabla\cdot\left(k(x,\omega)\nabla u(x,\omega)\right)=f(x,\omega),\]
and select the test function
\[h(x,c,r)=r^{-D_{x}}\cdot1_{\left\Vert x-c\right\Vert _{\infty}\le\frac{r}{2}},\]
where $r>0$ is a constant and $c$ is randomly chosen in $\mathcal D$ so that the support set of  $h(\cdot,c,r)$ is a subset of $\mathcal D$. It can be seen that $h(\cdot,c,r)$ defines a uniform distribution in the area $\{x|\left\Vert x-c\right\Vert _{\infty}\le\frac{r}{2}\}$.

The variational equation loss is then defined as
\begin{eqnarray*}
\mathcal{L}_{\mathrm{equ}} & = & \mathbb{E}_{\omega,c}\left[\left(-\nabla\cdot\left(k(\cdot,\omega)\nabla u(\cdot,\omega)\right)-f(\cdot,\omega),h(\cdot,c,r)\right)^{2}\right]\\
 & = & \mathbb{E}_{\omega,c}\left[\left(\left(k(\cdot,\omega)\nabla u(\cdot,\omega),\nabla h(\cdot,c,r)\right)-\left(f(\cdot,\omega),h(\cdot,c,r)\right)\right)^{2}\right],
\end{eqnarray*}
where the inner product $(f_1,f_2)\triangleq \int_{\mathcal D}f_1(x)^\top f_2(x)\mathrm dx$.

We now discuss how to obtain unbiased estimates of the variational equation loss for $D_x=1$ and $2$.

\subsubsection*{Case 1: $D_x=1$}
For this case,
\begin{eqnarray*}
\left(k(\cdot,\omega)\nabla u(\cdot,\omega),\nabla h(\cdot,c,r)\right) & = & r^{-1}\left(k(c-\frac{r}{2},\omega)\nabla u(c-\frac{r}{2},\omega)-k(c+\frac{r}{2},\omega)\nabla u(c+\frac{r}{2},\omega)\right),\\
\left(f(\cdot,\omega),h(\cdot,c,r)\right) & = & \mathbb{E}_{x\sim h(\cdot,c,r)}\left[f(x,\omega)\right].
\end{eqnarray*}
Therefore, we can obtain the unbiased estimate of $\mathcal{L}_{\mathrm{equ}}$ by implementing the following steps:
\begin{enumerate}
\item Sample $\xi_{1:n},\epsilon_{k,1:n},\epsilon_{f,1:n},\epsilon_{u,1:n}$ and $c_{1:n}$ from their prior distributions
\item Draw $x_i,x^\prime_i\sim h(\cdot,c_i,r_i)$ for $i=1,\ldots,n$
\item Calculate
\begin{eqnarray*}
\left(k(\cdot,\xi_{i},\epsilon_{k,i})\nabla u(\cdot,\xi_{i},\epsilon_{u,i}),\nabla h(\cdot,c,r)\right) & = & r^{-1}k(c_{i}-\frac{r}{2},\xi_{i},\epsilon_{k,i})\nabla u(c-\frac{r}{2},\xi_{i},\epsilon_{u,i})\\
 &  & -r^{-1}k(c_{i}+\frac{r}{2},\xi_{i},\epsilon_{k,i})\nabla u(c+\frac{r}{2},\xi_{i},\epsilon_{u,i})
\end{eqnarray*}
for $i=1,\ldots,n$.
\item Calculate
\[
\hat{\mathcal L}_{\mathrm{equ}}=\frac{1}{n}\sum_{i=1}^n \mathcal{L}_{\mathrm{equ},i}
\]
with
\begin{eqnarray*}
\mathcal{L}_{\mathrm{equ},i} & = & \left(\left(k(\cdot,\xi_{i},\epsilon_{k,i})\nabla u(\cdot,\xi_{i},\epsilon_{u,i}),\nabla h(\cdot,c,r)\right)-f(x_{i},\xi_{i},\epsilon_{f,i})\right)\\
 &  & \cdot\left(\left(k(\cdot,\xi_{i},\epsilon_{k,i})\nabla u(\cdot,\xi_{i},\epsilon_{u,i}),\nabla h(\cdot,c,r)\right)-f(x_{i}^{\prime},\xi_{i},\epsilon_{f,i})\right)
\end{eqnarray*}
\end{enumerate}

\subsubsection*{Case 2: $D_x=2$}
For the sake of convenience, we let $x=(a,b)^\top$ and $c=(\alpha,\beta)^\top$. We have
\[
\nabla h(a,b,c,r)=\left(\begin{array}{c}
r^{-2}\left(\delta\left(a-\alpha+\frac{r}{2}\right)-\delta\left(a-\alpha-\frac{r}{2}\right)\right)1_{\left|b-\beta\right|\le\frac{r}{2}}\\
r^{-2}\left(\delta\left(b-\beta+\frac{r}{2}\right)-\delta\left(b-\beta-\frac{r}{2}\right)\right)1_{\left|a-\alpha\right|\le\frac{r}{2}}
\end{array}\right)
\]
and
\begin{eqnarray*}
\left(k(\cdot,\omega)\nabla u(\cdot,\omega),\nabla h(\cdot,c,r)\right) & = & \iint k(a,b,\omega)\nabla_{a}u(a,b,\omega)\frac{\delta\left(a-\alpha+\frac{r}{2}\right)-\delta\left(a-\alpha-\frac{r}{2}\right)}{r^{2}}1_{\left|b-\beta\right|\le\frac{r}{2}}\mathrm{d}a\mathrm{d}b\\
 &  & +\iint k(a,b,\omega)\nabla_{b}u(a,b,\omega)\frac{\delta\left(b-\beta+\frac{r}{2}\right)-\delta\left(b-\beta-\frac{r}{2}\right)}{r^{2}}1_{\left|a-\alpha\right|\le\frac{r}{2}}\mathrm{d}a\mathrm{d}b\\
 & = & \mathbb{E}_{b\sim\mathcal{U}(\beta-\frac{r}{2},\beta+\frac{r}{2})}\left[\frac{k(\alpha-\frac{r}{2},b,\omega)\nabla_{a}u(\alpha-\frac{r}{2},b,\omega)-k(\alpha+\frac{r}{2},b,\omega)\nabla_{a}u(\alpha+\frac{r}{2},b,\omega)}{r}\right]\\
 &  & +\mathbb{E}_{a\sim\mathcal{U}(\alpha-\frac{r}{2},\alpha+\frac{r}{2})}\left[\frac{k(a,\beta-\frac{r}{2},\omega)\nabla_{b}u(a,\beta-\frac{r}{2},\omega)-k(a,\beta+\frac{r}{2},\omega)\nabla_{b}u(a,\beta+\frac{r}{2},\omega)}{r}\right].
\end{eqnarray*}
Then, the unbiased estimate of $\mathcal{L}_{\mathrm{equ}}$ can be obtained by the following steps:
\begin{enumerate}
\item Sample $\xi_{1:n},\epsilon_{k,1:n},\epsilon_{f,1:n},\epsilon_{u,1:n}$ and $\alpha_{1:n},\beta_{1:n}$ from their prior distributions
\item Draw $a_i,a^\prime_i\sim \mathcal U(\alpha-\frac{r}{2},\alpha+\frac{r}{2})$ and $b_i,b^\prime_i\sim \mathcal U(\beta-\frac{r}{2},\beta+\frac{r}{2})$ for $i=1,\ldots,n$
\item Calculate
\begin{eqnarray*}
e_{i} & = & r^{-1}k(\alpha_{i}-\frac{r}{2},b_{i},\xi_{i},\epsilon_{k,i})\nabla_{a}u(\alpha_{i}-\frac{r}{2},b_{i},\xi_{i},\epsilon_{u,i})\\
 &  & -r^{-1}k(\alpha_{i}+\frac{r}{2},b_{i},\xi_{i},\epsilon_{k,i})\nabla_{a}u(\alpha_{i}+\frac{r}{2},b_{i},\xi_{i},\epsilon_{u,i})\\
 &  & +r^{-1}k(a_{i},\beta_{i}-\frac{r}{2},\xi_{i},\epsilon_{k,i})\nabla_{b}u(a_{i},\beta_{i}-\frac{r}{2},\xi_{i},\epsilon_{u,i})\\
 &  & -r^{-1}k(a_{i},\beta_{i}+\frac{r}{2},\xi_{i},\epsilon_{k,i})\nabla_{b}u(a_{i},\beta_{i}+\frac{r}{2},\xi_{i},\epsilon_{u,i})\\
 &  & -f(a_{i},b_{i},\xi_{i},\epsilon_{f,i})\\
e_{i}^{\prime} & = & r^{-1}k(\alpha_{i}-\frac{r}{2},b_{i}^{\prime},\xi_{i},\epsilon_{k,i})\nabla_{a}u(\alpha_{i}-\frac{r}{2},b_{i}^{\prime},\xi_{i},\epsilon_{u,i})\\
 &  & -r^{-1}k(\alpha_{i}+\frac{r}{2},b_{i}^{\prime},\xi_{i},\epsilon_{k,i})\nabla_{a}u(\alpha_{i}+\frac{r}{2},b_{i}^{\prime},\xi_{i},\epsilon_{u,i})\\
 &  & +r^{-1}k(a_{i}^{\prime},\beta_{i}-\frac{r}{2},\xi_{i},\epsilon_{k,i})\nabla_{b}u(a_{i}^{\prime},\beta_{i}-\frac{r}{2},\xi_{i},\epsilon_{u,i})\\
 &  & -r^{-1}k(a_{i}^{\prime},\beta_{i}+\frac{r}{2},\xi_{i},\epsilon_{k,i})\nabla_{b}u(a_{i}^{\prime},\beta_{i}+\frac{r}{2},\xi_{i},\epsilon_{u,i})\\
 &  & -f(a_{i}^{\prime},b_{i}^{\prime},\xi_{i},\epsilon_{f,i})
\end{eqnarray*}
for $i=1,\ldots,n$.
\item Calculate
\[
\hat{\mathcal{L}}_{\mathrm{equ}}=-\frac{1}{n}\sum_{i=1}^{n}e_{i}e_{i}^{\prime}.
\]
\end{enumerate}

\end{document}